\pdfoutput=1
\documentclass[11pt]{article}
\usepackage[final]{acl}
\hypersetup{pdftitle={From Zero to Hero: An Open LLM Ecosystem for Armenian},
  pdfauthor={Erik Arakelyan, Khatun Avetisyan, Meri Davtyan, Heghine Grigoryan, Nane Khachatryan, Hayk Shahsuvaryan, Henrik Sergoyan, Vahan Martirosyan}}
\makeatletter
\ifdefined\@LN@makecol
  \AtBeginDocument{%
    \RemoveFromHook{build/column/before}[lineno]%
    \def\@LN@makecol{%
       \@LN@orig@makecol
       \setbox\@outputbox \vbox{%
          \boxmaxdepth \@maxdepth
          \protected@write\@auxout{}{%
             \string\def\string\@LN@column{\if@firstcolumn1\else2\fi}%
          }%
          \box\@outputbox
       }%
    }%
    \if@twocolumn \let\@makecol\@LN@makecol \fi
  }
\fi
\makeatother
\usepackage{times}
\usepackage{latexsym}
\usepackage{graphicx}
\usepackage{booktabs}
\usepackage{amsmath}
\usepackage{multirow}
\usepackage{xcolor}
\definecolor{eqcap}{RGB}{0,90,160}
\definecolor{eqirr}{RGB}{170,85,0}
\newcommand{\armweb}{ArmWeb}
\newcommand{\armstem}{ArmSTEM}
\newcommand{\ourmodel}{arm-gemma-e4b}
\newcommand{\newsrun}{News-CPT}
\newcommand{\stemrun}{STEM-CPT}
\newcommand{\fullrun}{STEM-CPT-full}

\title{From Zero to Hero: An Open LLM Ecosystem for Armenian}

\author{Erik Arakelyan$^{1,2}$ \quad Khatun Avetisyan$^{2}$ \quad
  Meri Davtyan$^{2}$ \quad Heghine Grigoryan$^{2}$ \\
  \textbf{Nane Khachatryan$^{2}$ \quad Hayk Shahsuvaryan$^{2}$ \quad
  Henrik Sergoyan$^{2}$ \quad Vahan Martirosyan$^{2}$} \\
  $^{1}$NVIDIA \quad $^{2}$COPA \\
  \texttt{earakelyan@nvidia.com}, \texttt{info@copa.team}}

\begin{document}
\maketitle

\begin{abstract}
Pretraining data for Armenian, a morphologically rich and low-resource
language, is scarce, and no open Armenian LLM has been released with the
data and recipe needed to reproduce it. To address this gap, we curate and
release two datasets. \armweb{} is an extensively validated corpus of
4.37M Armenian news documents. \armstem{} is a parallel English--Armenian
collection of 373K math and science problems, 324K of them with
step-by-step solutions,
translated into Armenian and verified through both answer-preserving LLM
judgment and human evaluation. Continued pretraining of Gemma-4-E4B on these datasets yields
\ourmodel, which outperforms every existing open Armenian model as well as
its unadapted base, and is the first open Armenian LLM with complete
training data and recipe. Our ablations show that news-only continued
pretraining improves fluency while eroding knowledge, a pattern we also
observe in existing Armenian models, and that a small share of verified
translated STEM data reverses the loss. We further find that the largest public Armenian corpora overlap
web-derived evaluation panels heavily, including a train/test self-overlap
inside FineWeb-2. We
openly release all data, models, and code.
\end{abstract}

\section{Introduction}

Progress in language modeling remains concentrated in English and a
handful of data-rich languages \citep{bloom2022,aya2024}. Although
massively multilingual models train on corpora spanning hundreds of
languages, their benefits are uneven, since pretraining allocates most
capacity to the same high-resource head of the language distribution
\citep{bloom2022}. Small \emph{monolingual} models still outperform models
orders of magnitude larger on basic generation in low-resource languages
\citep{goldfish2026}. To overcome this imbalance, a growing line of work
releases language-specific ecosystems, publishing a curated corpus, an
adapted open model, evaluations, and the training recipe together, as has
been done for Basque \citep{latxa2024}, Kazakh \citep{sherkala2025},
Polish \citep{pllum2025}, and Southeast Asian languages
\citep{seallms2024}.

Armenian is a morphologically rich language that is widely considered
low-resource \citep{syndarin2025,eanc2022}, and it has no such
ecosystem.
Its open text is confined to Armenian slices of multilingual web crawls
\citep{oscar2022,penedo2025fineweb2,hplt2025} with no Armenian-specific
curation, and we show below that these slices overlap Armenian
evaluation sets at rates up to 17.4\% (\S\ref{sec:armweb}), echoing audits of
English corpora \citep{dodge2021c4audit,sainz2023contamination}. Its open models are
released as weights alone, without training data or recipes
\citep{hygpt2025,remy2024transtok}, so they cannot be audited or
reproduced. Evaluation, meanwhile, has recently improved
\citep{armbench2025,syndarin2025}, but there is no open training data
for the knowledge those benchmarks measure. No math or science text with
worked solutions exists in Armenian at training scale.

This paper releases the missing pieces as one documented, auditable
pipeline,\footnote{Data and model: \url{https://huggingface.co/COPA-AI}.
Code: \url{https://github.com/COPATeam/armenian_llm_ecosystem}.} following the datasheet practice of \citet{gebru2021datasheets}
and the per-stage reporting of \citet{dolma2024}, and reports a finding
about language adaptation that we believe generalizes. We make five
contributions.

\begin{figure}[t]
\centering
\includegraphics[width=0.82\columnwidth]{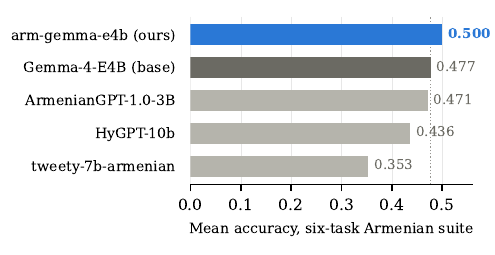}
\caption{Mean accuracy on the six-task Armenian likelihood suite
(\S\ref{sec:setup}) for all open Armenian models and the unadapted
base.}
\label{fig:headline}
\end{figure}

\begin{itemize}
\item We release \textbf{\armweb}, a 4.37M-document (3.3B Gemma-token)
Armenian news corpus built from an author-operated 15-year crawl with a
fully documented pipeline covering extraction, language identification,
syndication-aware deduplication, verified leakage gates, and 13-gram
decontamination against ten Armenian benchmarks, with per-document
provenance metadata. Quality is validated by a 410M ablation grid, a
repetition study, and a scaling ladder whose fit predicts the loss of a
held-out 1.3B model to within 0.47\%.
\item We release \textbf{\armstem}, 373K mathematics and science
problems from GSM8K \citep{gsm8k2021}, AceReason-Math
\citep{acereason2025}, OpenScience \citep{openscience2025}, and
OpenScienceReasoning-2 \citep{osr2_2025},
machine-translated to Eastern Armenian and \emph{verified} through
placeholder-masked translation, language identification, and blind
re-solving with exact-match answer checking, with dual-side
decontamination against both English benchmark origins and Armenian
benchmark targets. In a human evaluation, two native speakers
independently rated 299 of 300 sampled problems as valid, with perfect
inter-annotator agreement. The corpus ships as parallel EN--HY data with per-item provenance and
per-subset licenses, and it is to our knowledge the first Armenian STEM
corpus with step-by-step solutions at training scale.
\item We release \textbf{\ourmodel}, Gemma-4-E4B adapted to Armenian
by continued pretraining on a 69/6/20/5 mixture of \armweb, \armstem{}
(both languages), English web replay, and code. It outperforms every open Armenian model we are aware of, as well as
its own unadapted base (Figure~\ref{fig:headline}), and it is to our knowledge the first open
Armenian LLM released with its complete training corpus and recipe.
\item We provide \textbf{a rigorous study of the adaptation recipe}. Naive
continued pretraining on news text exhibits classic catastrophic
forgetting \citep{mccloskey1989forgetting,french1999forgetting}, trading
knowledge for fluency at up to $-21.2$pp on Belebele
\citep{belebele2024}, and existing
Armenian models show the same pattern of improved fluency and degraded
knowledge at a comparable budget. A gentler learning rate
\citep{ibrahim2024cpt} recovers roughly two-thirds of the loss, and
epoch-capped verified translated STEM data more than reverses it, ending
$+2.2$pp \emph{above} the base model on average while keeping the fluency
gains (\S\ref{sec:results}).
\item We report \textbf{benchmark hygiene results}. Scans of the three
largest public Armenian crawl slices find 7.9--17.4\% of documents
overlapping our evaluation sets, dominated by web-derived perplexity
panels and including FineWeb-2's own Armenian test split leaking into its
training split. Without decontamination, perplexity-style evaluation
systematically overstates the performance of crawl-trained Armenian
models.
\end{itemize}

Figure~\ref{fig:headline} summarizes the central result.

\section{Methodology}
\label{sec:method}

\paragraph{Token units.} This paper counts tokens in two tokenizers.
\emph{Gemma tokens} use the stock Gemma-4 vocabulary (4.15 tokens per
Armenian word, Table~\ref{tab:fertility}), the unit the released model was
trained in, so all corpus sizes, mixture shares, and cross-corpus
comparisons use them, with other public corpora re-tokenized by us.
\emph{SP tokens} use a 32K SentencePiece tokenizer trained on \armweb{},
the most fertile tokenizer we built (1.48 tokens per word). The small-scale
Megatron ablations of \S\ref{sec:armweb} use it so that small models spend
their budget on content rather than on fragmenting words, so their budgets
are stated in SP tokens. One SP token is about 2.8 Gemma tokens on
\armweb{} text.

We build a curated in-language corpus (\S\ref{sec:armweb}) and a
verified translated knowledge corpus (\S\ref{sec:armstem}), which
\S\ref{sec:model} combines into an adaptation recipe.

\subsection{The \armweb{} news corpus}
\label{sec:armweb}

\paragraph{Source.} \armweb{} derives from a single-operator crawl of
public Armenian news sites spanning 2011--2026, stored as a structured
database rather than raw HTML, which nearly eliminates the boilerplate
that dominates web-crawl curation. Each record carries title, body, URL,
outlet, topic, and publication and crawl dates. We release all metadata
except author names.

\paragraph{Pipeline.} Table~\ref{tab:funnel} gives the full funnel.
Cleaning is deliberately minimal because the heuristics that dominate
web-crawl curation, such as boilerplate rules, blocklists, and quality
classifiers \citep{raffel2020c4,rae2021gopher,refinedweb2023}, target
crawl artifacts our source lacks and are known to discard benign text as
collateral \citep{dodge2021c4audit}. Following the light-touch precedent
of OSCAR \citep{oscar2022}, we apply only title--body concatenation, a
minimum length of 100 characters, and whitespace normalization. The
released text is NFC-normalized and otherwise verbatim, while all
similarity computations use a separate aggressive ``signature view'' built
with NFKC normalization, Armenian ligature folding, punctuation stripping,
and digit-zeroing. GlotLID \citep{glotlid2023} retains 98.3\% of documents as
\texttt{hye/hyw}. Deduplication measurably improves the
resulting models \citep{lee2022dedup,refinedweb2023,dclm2024} and reduces
memorization \citep{kandpal2022privacy}, and it must run globally
\emph{before} splitting because duplicates that straddle a train/test
boundary silently inflate evaluation \citep{lee2022dedup,slimpajama2023}.
We first remove exact duplicates with xxh128 hashing and a keep-longest
rule, then run MinHash LSH with word 5-gram shingles, 112 permutations,
and 14 bands of 8 rows, targeting a Jaccard threshold near 0.72. The low
threshold is our tuned choice, since news syndication produces true
reprints at far lower Jaccard similarity than web duplicates
\citep{silcock2023newsdedup} and standard 0.8-threshold recipes
\citep{refinedweb2023} miss most of them, an instance of the per-source
tuning FineWeb2 argues is necessary \citep{penedo2025fineweb2}.
Appendix~\ref{app:dedupcheck} validates the engine against two independent
implementations. Dolma-style repeated paragraph removal
\citep{dolma2024} modifies 0.1\% of documents and drops only the 13 that
fall below the length floor afterward.

\paragraph{Splits and leakage checks.} We hold out three evaluation
splits of 20K documents each. The validation and in-distribution test
splits are drawn by stratified sampling over outlet and month, so each
mirrors the training distribution, while the temporal test split takes
documents from the final two months of the crawl to measure
generalization to future text. Because deduplication ran before
splitting, no held-out document should have a duplicate in training, and
three checks verify this. The first requires zero exact-duplicate
documents between any two splits. The second bounds near-duplicates,
found with the same MinHash procedure, at 0.1\% of held-out documents.
Both run as hard assertions that abort the release build if violated,
and both pass, with zero exact collisions and near-duplicate rates of
0.045\%, 0.010\%, and 0.005\% for validation, in-distribution test, and
temporal test. The third check counts paragraphs of at least 13 tokens
that a held-out split shares with training and reports the counts rather
than asserting a bound, finding 30, 63, and 10 shared paragraphs for the
three splits (Appendix~\ref{app:datasheet}). Finally, training text
is decontaminated by 13-gram overlap against ten Armenian evaluation sets,
removing 147{,}101 documents (3.3\%), following the n-gram
decontamination practice operationalized by open-corpus tooling
\citep{dolma2024,dclm2024}.

\begin{table}[t]\centering\small
\begin{tabular}{lrr}
\toprule
Stage & Documents & $\Delta$ \\
\midrule
Extracted & 5{,}918{,}811 & --- \\
Language ID & 5{,}817{,}343 & $-1.7\%$ \\
Exact dedup & 5{,}436{,}984 & $-6.5\%$ \\
MinHash dedup & 4{,}515{,}497 & $-16.9\%$ \\
Boilerplate & 4{,}515{,}484 & $-0.0\%$ \\
Splits held out & 4{,}455{,}484 & $-60$K \\
Decontamination & 4{,}308{,}383 & $-3.3\%$ \\
\midrule
\multicolumn{3}{l}{Final: 11.0\,GB, 3.3B Gemma / 1.15B SP tokens} \\
\bottomrule
\end{tabular}
\caption{\armweb{} pipeline funnel. Each $\Delta$ is relative to the
preceding row. ``Splits held out'': val/test 20K each plus a 20K temporal
tail, removed before train-side decontamination.}
\label{tab:funnel}
\end{table}

\paragraph{Contamination of existing corpora.} Benchmark contamination is
documented in English corpora \citep{dodge2021c4audit}, invalidates
evaluation conclusions when unreported \citep{sainz2023contamination}, and
is detectable post hoc \citep{oren2024contamination}, so it will
eventually be audited. Applying our scan to the Armenian slices of the three largest recent
public corpora quantifies a problem the community has not priced in. Each
overlaps our ten evaluation sets at document rates of 7.9--17.4\%
(Table~\ref{tab:contamination}), dominated by the web-derived perplexity
panels, above all FineWeb-2's own Armenian test split and Wikipedia, while
hits on the hand-built MCQA benchmarks are near zero for every corpus
including ours. Restricted to the hand-built benchmarks, document rates are at or below
0.06\% for every corpus, though only 20--53\% of Belebele
\citep{belebele2024}, INCLUDE \citep{include2025}, and SynDARin
\citep{syndarin2025} items are long enough to be detectable at
$n{=}13$, and
FineWeb-2 additionally carries 38.8K documents overlapping ArmBench items.
Perplexity-style evaluation of crawl-trained Armenian models is therefore
inflated, FineWeb-2's own train/test split leaks, and our bpb comparisons
(Table~\ref{tab:grid}) are fair only because these panels were
decontaminated from \armweb.

\begin{table}[t]\centering\small
\setlength{\tabcolsep}{4pt}
\resizebox{\columnwidth}{!}{%
\begin{tabular}{lrrr}
\toprule
Corpus & Docs & Gemma tokens & Contaminated \\
\midrule
\textbf{\armweb{} (ours)} & 4.46M & 3.3B & 3.3\% \\
CulturaX-hy & 2.96M & 4.5B & 7.9\% \\
HPLT-v2-hy & 3.60M & 5.8B & 10.9\% \\
FineWeb-2-hy & 1.76M & 2.3B & 17.4\% \\
\bottomrule
\end{tabular}}
\caption{Documents sharing at least one 13-gram with our ten Armenian
evaluation sets (\S\ref{sec:setup}), over the normalized signature view.
The \armweb{} row is the \emph{pre-removal} training pool, and released
splits have all hits removed. Rates are document-level and
length-sensitive. Gemma tokens re-tokenize each corpus as distributed
with the Gemma-4 vocabulary (\armweb{} as released), so sizes are
comparable across corpora.}
\label{tab:contamination}
\end{table}

\paragraph{Is curation worth it?} We answer with controlled small-scale
ablations, relying on the finding that data-recipe rankings established at
small scale predict rankings at larger scale \citep{datadecide2025,dclm2024}.
In 410M-parameter comparisons with identical architecture, tokenizer,
and a 4.6B SP-token budget (Table~\ref{tab:grid}), spanning CulturaX
\citep{culturax2023}, HPLT-v2, FineWeb-2, and news (LR-Sum
\citep{lrsum2023}), wiki, and FLORES \citep{nllb2024} panels, \armweb{}
beats all crawl slices on held-out news by roughly 10\% bits-per-byte,
while crawls win on web and wiki text. The \emph{union} of \armweb{} and CulturaX wins on the mean, with a
bpb of 0.532 against 0.545 for the best single corpus, and is within
0.007 bpb of the column best on every panel. Because the crawl trainers
retain their contamination on the web-derived panels
(Table~\ref{tab:contamination}), we also check the three panels on which
no trainer has a single 13-gram hit (our held-out news pair and FLORES),
where the union still leads at 0.525 against 0.544 for \armweb{} and
0.548 for CulturaX, so the conclusion is not a train-on-test artifact.
Mixing Russian into the training stream instead degrades Armenian bpb
monotonically at fixed compute (Table~\ref{tab:grid}). The ranking transfers exactly to 1.3B confirmation runs, where the
union's lead \emph{widens} (Appendix~\ref{app:ablations}). Curated and
crawled Armenian data are complements, not substitutes.

\begin{table*}[t]\centering\small
\begin{tabular}{lccccccc}
\toprule
Variant & our-iid & our-tail & FW2-test & hyWiki & LR-Sum & FLORES & Avg\\
\midrule
\textbf{Union (\armweb+CulturaX)} & 0.426 & 0.433 & 0.545 & \textbf{0.603} & \textbf{0.468} & 0.717 & \textbf{0.532}\\
CulturaX-hy & 0.465 & 0.469 & \textbf{0.543} & 0.610 & 0.472 & \textbf{0.710} & 0.545\\
FineWeb-2-hy & 0.482 & 0.472 & 0.538 & 0.643 & 0.477 & 0.735 & 0.558\\
\armweb\ (ours) & \textbf{0.419} & \textbf{0.426} & 0.589 & 0.666 & 0.477 & 0.788 & 0.561\\
HPLT-v2-hy & 0.474 & 0.483 & 0.584 & 0.620 & 0.477 & 0.737 & 0.562\\
ArmWeb/Ru 90/10 & 0.424 & 0.429 & 0.593 & 0.660 & 0.480 & 0.790 & 0.563\\
ArmWeb/Ru 75/25 & 0.428 & 0.433 & 0.596 & 0.665 & 0.484 & 0.790 & 0.566\\
ArmWeb/Ru 50/50 & 0.439 & 0.444 & 0.608 & 0.679 & 0.493 & 0.798 & 0.577\\
\bottomrule
\end{tabular}
\caption{Cross-corpus bits-per-byte at 410M/4.6B SP tokens (lower is
better). Bold marks the column best, excluding FineWeb-2 on its own test
panel. Crawl trainers are used as distributed and keep their contamination
on the FW2-test and hyWiki panels (see text). ArmWeb/Ru rows mix
\armweb{} with a Russian sister collection at the stated ratios.}
\label{tab:grid}
\end{table*}

\paragraph{Effect of data repetition.} At 1.15B SP tokens, \armweb{} is
small enough that any realistic training run will repeat it, so we
measure the value of repetition directly. The marginal gain shrinks with each doubling of epochs at 160M
and stays positive through 8 epochs at both scales with no saturation
cliff (Table~\ref{tab:epochs}), consistent with
data-constrained scaling laws
\citep{muennighoff2023dataconstrained,hernandez2022repetition}. This
trend directly validates the epoch-capped mixture of \S\ref{sec:model}.
The repetition runs were trained separately from the ablation grid, and
the 410M 4-epoch run scores 0.559 mean bpb against the grid's \armweb{}
score of 0.561, independently replicating the grid measurement.

\begin{table}[t]\centering\small
\setlength{\tabcolsep}{5pt}
\begin{tabular}{lcccc}
\toprule
Epochs over \armweb & 1 & 2 & 4 & 8\\
\midrule
160M mean bpb & 0.710 & 0.628 & 0.593 & 0.572\\
410M mean bpb & 0.649 & 0.590 & 0.559 & 0.526\\
\bottomrule
\end{tabular}
\caption{Mean panel bpb when \armweb{} is repeated for more epochs, at
two model scales (lower is better).}
\label{tab:epochs}
\end{table}

\begin{figure}[t]\centering
\includegraphics[width=0.82\columnwidth]{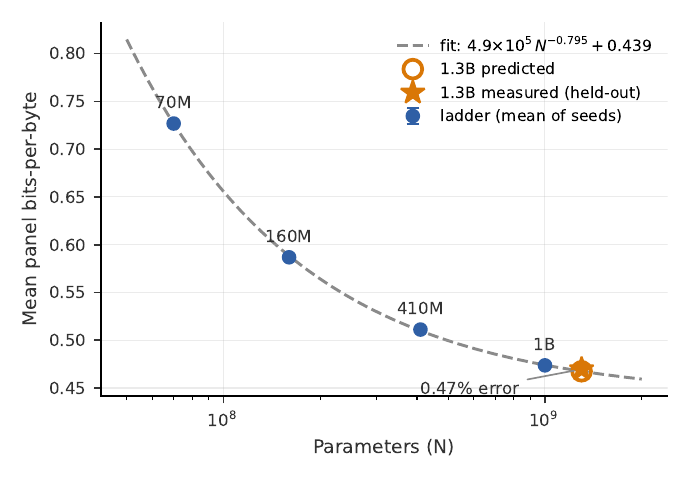}
\caption{Scaling ladder on the union recipe. Points are Chinchilla-budget
runs from 70M to 1B, the curve is the power-law fit, and the held-out
point is the independently trained 1.3B confirmation. Seed-range bars
(3/3/3/2 per rung) are smaller than the markers.}
\label{fig:ladder}
\end{figure}

\paragraph{Scaling ladder.} The ablations above are measured at 160M
and 410M parameters, so their value depends on whether conclusions drawn
at that scale carry over to larger models. The scaling ladder tests this
directly. We train the union recipe at four compute-optimal budgets
\citep{hoffmann2022chinchilla}, from 70M parameters at 1.4B SP tokens to 1B
at 20B, with 3/3/3/2 seeds per rung and seed spreads at or below 0.003
bpb. Mean panel bpb falls from 0.727 through 0.587 and 0.511 to 0.474
across the ladder. Residual duplication or contamination that a larger model could exploit
would bend the curve away from the smooth power law that clean
pretraining data produces \citep{kaplan2020scaling}. Instead a
three-parameter power law (capacity scale, exponent, and irreducible
floor) fits all four rungs within 0.0015 bpb at full precision and
extrapolates to a fifth, independently trained, larger model. A power-law fit of mean panel bpb against parameter count
$N$,
\begin{equation}
\label{eq:ladder}
\mathrm{bpb}(N)=
\textcolor{eqcap}{\underbrace{\textcolor{black}{4.9{\times}10^{5}\,N^{-0.795}}}_{\text{\footnotesize capacity term }A\,N^{-\alpha}}}
+
\textcolor{eqirr}{\underbrace{\textcolor{black}{0.439}}_{\text{\footnotesize irreducible bpb }E}}
\end{equation}
predicts $0.4675$ for the \emph{held-out} 1.3B confirmation model
against a measured $0.4697$, a \textbf{0.47\% extrapolation error} on a
model 30\% larger than the largest model in the fit and trained
independently (Figure~\ref{fig:ladder}). The irreducible term $E$
estimates the bpb floor of the panel under this recipe, and the exponent
$\alpha$ measures how quickly added capacity buys loss. Together with the
1.3B confirmation runs, which preserve the 410M corpus ranking
(Appendix~\ref{app:ablations}), the ladder shows that the grid's
conclusions are properties of the data rather than of the 410M scale.
Coefficients are rounded for display and the full-precision fit ships
with the code.

\subsection{The \armstem{} translated STEM corpus}
\label{sec:armstem}

Translated reasoning data reliably improves target-language ability.
Chain-of-thought transfers across languages \citep{mgsm2023}, translated
math training data improves multilingual math performance
\citep{mathoctopus2024}, and translated reasoning traces can even beat
native-language distillation from stronger models
\citep{barua2025translatedcot}. Armenian cannot exploit any of this because it has no math or science
training data with worked solutions, only small evaluation sets
\citep{include2025,armbench2025}. \armstem{} fills this gap by
translating verified English corpora, namely GSM8K \citep{gsm8k2021},
AceReason-Math \citep{acereason2025}, OpenScience \citep{openscience2025},
and OpenScienceReasoning-2 \citep{osr2_2025}, selected for
machine-checkable answers, permissive licenses, and freedom from benchmark
provenance.

\paragraph{Translation with placeholder masking.} The dominant failure mode
of translating math is corruption of numbers and notation
\citep{mathoctopus2024}. We eliminate it structurally rather than
detecting it after the fact, adapting the placeholder/do-not-translate
masking long used for numerals, entities, and inline markup in NMT
\citep{crego2016systran,postvilar2018,dinu2019terminology} to
LaTeX-bearing STEM text. Numbers, LaTeX spans, and a question/solution
separator are replaced by indexed placeholder tokens before translation
and restored afterward. We translate with Gemini-3.1-flash-lite
\citep{gemini31_2026}. To choose it, we translated the same 200 items with
Gemini-3.1-flash-lite and with GPT-5.5 \citep{gpt55_2026} and passed both
sets of outputs through the full verification pipeline described next.
The decision rule was fixed before scoring, namely that we would switch
to the costlier GPT-5.5 only if its pass rate exceeded Gemini's by at
least 5 points. GPT-5.5 passed 96.5\% of the items and Gemini 94.0\%, a
2.5-point gap below the threshold, so we kept the cheaper model.

\paragraph{Verification gates.} Where LLM-based translation evaluation
typically judges fluency and adequacy \citep{gemba2023}, STEM data admits
a stronger, \emph{functional} check, namely whether the translated problem
still has the same answer. Each item passes three gates in order. Gate G0 checks placeholder
integrity, requiring every token exactly once and no stray digits. Gate G1
is a GlotLID language check. Gate G2 is \emph{blind re-solving}, in which
an independent model (o4-mini, \citealp{o4mini2025}) solves the Armenian problem and must
reproduce the gold answer exactly. On a mismatch, a control re-solves the
English original, and if that also fails the item is solver-limited rather
than mistranslated and is kept with a tag, which affects 10.9\% of
accepted math and 27.6\% of science items and certifies problem-statement
integrity only. Freeform answers, under 5\% of items, use a three-model
judge panel with 2/3 majority \citep{gemba2023}. Failures are retried
twice with feedback, then escalated to a stronger translator, then
dropped. Per-source acceptance after repair ranges from 91.5\% (OSR-2) to
99.1\% (GSM8K), 96.6\% overall (Table~\ref{tab:armstem}), and per-stage
rejection statistics ship with the corpus. An automated adequacy audit (GPT-5.5 judging 300 stratified items
against their sources, 1--5 scale) rates 100\% of re-solve-verified and
92.7\% of solver-limited items meaning-preserving (means 4.65 and 4.39),
bounding the tagged stratum's residual risk. The adequacy judge also
serves as the escalation translator for a small fraction of items. For a
complete verification of the translated samples, two native Armenian
speakers independently assessed a sample of 300 \armstem{} problems,
judging whether each Armenian problem makes logical sense and whether its
solution is correct, and rated 299 of 300 as valid, with identical
verdicts on every item (raw agreement 100\%, Cohen's $\kappa=1.0$). Answer-checking has
precedent for
\emph{generated} math data \citep{openmathinstruct2_2024}. Applying it to
\emph{translation} is, to our knowledge, new. G2 checks answer
preservation rather than solution style (\S\ref{sec:limitations}).

\paragraph{Dual-side decontamination.} Because Armenian benchmarks are
partly translations of English ones, one scan cannot suffice. English
sources are scanned against English benchmark origins, above all the
MMLU-Pro test set that feeds ArmBench's largest column, and accepted
Armenian translations are scanned against the full Armenian benchmark item
set. Both gates rejected real collisions in production.
Table~\ref{tab:armstem} gives the resulting composition, totalling
372{,}907 pairs, about 311M Armenian and 124M parallel English Gemma
tokens. A harder competition-math tranche is in preparation.

\begin{table}[t]\centering\small
\begin{tabular}{lrrr}
\toprule
Source & Pool & Accepted & Rate \\
\midrule
GSM8K & 7{,}473 & 7{,}404 & 99.1\% \\
AceReason-Math & 49{,}585 & 48{,}584 & 98.0\% \\
OpenScience & 271{,}440 & 264{,}266 & 97.4\% \\
OSR-2 & 57{,}573 & 52{,}653 & 91.5\% \\
\midrule
Total & 386{,}071 & 372{,}907 & 96.6\% \\
\bottomrule
\end{tabular}
\caption{\armstem{} composition. ``Pool'' is the full English source.
``Accepted'' is the verified EN--HY pairs released. ``Rate'' is
Accepted/Pool.}
\label{tab:armstem}
\end{table}

\section{Adapting Gemma-4 to Armenian}
\label{sec:model}

Continued pretraining is the established route to language adaptation
when target-language text is too scarce to pretrain from scratch
\citep{gururangan2020dapt,cui2023chinesellama,sambalingo2024}. We adapt
Gemma-4-E4B, chosen among CPT candidates for the best Armenian tokenizer
fertility at 4.15 tokens per word against Qwen3.5's 5.31 and Llama-3.1's
12.2 (Table~\ref{tab:fertility}), since high fertility inflates both
compute cost and downstream error on the affected language
\citep{rust2021fertility,ahia2023tokencost}.

\paragraph{Keep or extend the tokenizer?} A natural alternative to living
with a multilingual tokenizer is extending its vocabulary with Armenian
tokens, which pays off at large token budgets \citep{cui2023chinesellama}
but is known to be sensitive to budget and initialization
\citep{yamaguchi2024vocab,zhao2024transfer}. We settle the question with a
controlled ablation (Table~\ref{tab:tokab}). \emph{Mean-initialized}
extension is sharply harmful at the 2B-token ablation budget, landing
40--45\% above even the unadapted base in bits-per-byte, as new embeddings displace well-trained BPE compositions
faster than they can be learned, so we retain the stock tokenizer. The
ablation uses the smaller E2B at a fifth of the final budget, and stronger
initializers \citep{wechsel2022,focus2023} remain untested, so extension
could become viable at larger scale.

\begin{table*}[t]\centering\small
\begin{tabular}{lccccccc}
\toprule
Model & our-iid & our-tail & FW2-test & hyWiki & LR-Sum & FLORES & Mean\\
\midrule
Base E2B (un-adapted) & 0.463 & 0.460 & 0.375 & 0.188 & 0.471 & 0.731 & 0.448\\
\textbf{CPT, stock tokenizer} & \textbf{0.356} & \textbf{0.378} & \textbf{0.325} & \textbf{0.174} & \textbf{0.404} & \textbf{0.684} & \textbf{0.387}\\
CPT, $+$8k hy tokens & 0.613 & 0.635 & 0.565 & 0.365 & 0.656 & 0.933 & 0.628\\
CPT, $+$16k hy tokens & 0.635 & 0.655 & 0.591 & 0.401 & 0.674 & 0.948 & 0.651\\
\bottomrule
\end{tabular}
\caption{CPT tokenizer ablation (Gemma-4-E2B, 2B tokens in each variant's
own tokenizer).
Bits-per-byte, lower is better. Because bpb is tokenizer-independent, the
comparison is fair across vocabularies.}
\label{tab:tokab}
\end{table*}

\paragraph{Mixture and schedule.} Training runs for 10B tokens,
sequence-packed, on a five-stream mixture of \textbf{69\% \armweb{}},
\textbf{4\% \armstem-HY}, \textbf{2\% \armstem-EN} (parallel data improves
both sides \citep{mathoctopus2024,transwebedu2024}), 20\% English web
replay from FineWeb-Edu \citep{fineweb2024}, the standard mitigation for
forgetting under distribution shift
\citep{rolnick2019replay,ibrahim2024cpt}, and 5\% code from Stack-smol
\citep{thestack2022}. We compare five training runs and name them by
their data. \newsrun{} trains on news only, with the same replay and code
streams at 75/20/5, at learning rates $10^{-4}$ and $3{\times}10^{-5}$.
\stemrun{} trains on the five-stream mixture above at the same two
learning rates, and \stemrun{} at $3{\times}10^{-5}$ is the released
\ourmodel. Crucially for attribution, \emph{the \newsrun{} runs use the
identical trainer, seed, budget, and replay and code streams}, so a
\stemrun{} run differs from its \newsrun{} counterpart only in swapping 6
points of \armweb{} for \armstem. The \armstem{} stream in these runs is a
109{,}885-item subset of the corpus, about 30\%, sampled at random with
balanced stratification across the mathematics and science pools (51/49
by items). The fifth run, \fullrun, trains the released recipe with its
STEM share drawn from the full 373K-item corpus instead, and
\S\ref{sec:results} shows it matches \ourmodel{} on the likelihood suite
while trailing it on format-sensitive generative tasks. Mixture weights are \emph{epoch-capped} because repetition beyond
roughly 4 epochs yields rapidly diminishing returns
\citep{muennighoff2023dataconstrained,hernandez2022repetition}. In
training-tokenizer accounting \armweb{} spans roughly 3.3B Gemma tokens,
so its share is read about twice, which is near-free, while each
\armstem{} token is read 7--9 times, within the productive zone measured
in Table~\ref{tab:epochs}. The learning rate of $3{\times}10^{-5}$
cosine, the single most consequential CPT hyperparameter
\citep{ibrahim2024cpt}, is itself an experimental result
(\S\ref{sec:results}), three times gentler than our first attempt at
$10^{-4}$.

\section{Experimental setup}
\label{sec:setup}

We evaluate on two complementary suites. The first is a six-task
Armenian likelihood suite comprising Belebele-hye \citep{belebele2024},
m-MMLU-hy \citep{mmlu2021}, INCLUDE-Armenian \citep{include2025}, ARC-hy
\citep{arc2018}, HellaSwag-hy \citep{hellaswag2019}, and MultiBLiMP-hye
\citep{multiblimp2026}, scored by zero-shot log-likelihood accuracy in LM
Evaluation Harness \citep{evalharness2024}, which measures knowledge
independent of output formatting. Item counts range from 550 (INCLUDE) to 10{,}891 (m-MMLU), and the -hy
versions of m-MMLU, ARC, and HellaSwag are Okapi machine translations
\citep{okapi2023}, a provenance we account for in \S\ref{sec:results}.
The second is ArmBench-LLM \citep{armbench2025}, a 24-task generative
benchmark including Armenian national-exam sections and MMLU-Pro-Hy
\citep{mmlupro2024}, run in the authors' lighteval \citep{lighteval2023}
fork (Appendix~\ref{app:armbench}). In
the main text we report the ArmBench tasks whose metrics measure knowledge
and language competence for \emph{base} models. The remaining tasks score
instruction adherence and output formatting, which base models
definitionally lack, so we report them in the appendix and defer them to
instruction-tuned variants. The ten \armweb{} decontamination targets are Belebele, INCLUDE,
HellaSwag-hy, MultiBLiMP, SIB-200, SynDARin, LR-Sum, the FineWeb-2-hy test
split, hyWiki eval, and FLORES-200 \citep{nllb2024}, and the CPT mixture
was additionally scanned against all ArmBench items including MMLU-Pro-Hy
(m-MMLU-hy and ARC-hy were not \armweb{} targets, so we scanned both post
hoc, finding zero of the 4.31M \armweb{} training documents and 4
of the 372{,}907 \armstem{} pairs sharing any 13-gram with their items). As baselines we
evaluate the unadapted Gemma-4-E4B, the existing open Armenian models
HyGPT-10b \citep{hygpt2025}, ArmenianGPT-1.0-3B \citep{armeniangpt2026}
(a Mistral-3-based instruction-tuned model, evaluated as described in
Appendix~\ref{app:repro}), and tweety-7b-armenian
\citep{remy2024transtok}, and finally Gemma-2-9B, which is HyGPT's own
base model and lets us measure that model's adaptation delta.

\section{Results}
\label{sec:results}

\begin{table}[t]\centering\small
\setlength{\tabcolsep}{3pt}
\resizebox{\columnwidth}{!}{%
\begin{tabular}{lcrrrrrr}
\toprule
 & & & \multicolumn{2}{c}{\newsrun} & \multicolumn{3}{c}{\stemrun} \\
\cmidrule(lr){4-5}\cmidrule(lr){6-8}
Task & {\scriptsize 95\%} & base & {\scriptsize$10^{-4}$} & {\scriptsize$3{\times}10^{-5}$} & {\scriptsize$10^{-4}$} & {\scriptsize$\mathbf{3{\times}10^{-5}}$} & {\scriptsize full} \\
\midrule
MultiBLiMP & {\scriptsize$\pm$.005} & 0.989 & \textbf{0.995} & 0.994 & \textbf{0.995} & 0.992 & 0.993 \\
HellaSwag-hy & {\scriptsize$\pm$.010} & \textbf{0.266} & 0.263 & 0.263 & 0.264 & 0.262 & 0.263 \\
ARC-hy & {\scriptsize$\pm$.025} & 0.227 & 0.203 & 0.216 & 0.217 & \textbf{0.229} & 0.219 \\
m-MMLU-hy & {\scriptsize$\pm$.009} & \textbf{0.343} & 0.272 & 0.310 & 0.314 & 0.337 & 0.334 \\
INCLUDE & {\scriptsize$\pm$.041} & 0.416 & 0.335 & 0.436 & 0.391 & \textbf{0.456} & 0.455 \\
Belebele & {\scriptsize$\pm$.032} & 0.619 & 0.407 & 0.550 & 0.590 & \textbf{0.716} & 0.703 \\
\midrule
Mean & {\scriptsize$\pm$.010} & 0.477 & 0.412 & 0.462 & 0.462 & \textbf{0.500} & 0.494 \\
\bottomrule
\end{tabular}}
\caption{The likelihood suite across the five adaptation runs of
\S\ref{sec:model}. Accuracy, with the row best in bold. The second column
gives per-task 95\% binomial half-widths. The bold learning rate marks the
released \ourmodel{} and ``full'' is \fullrun. Each column is a single
training run.}
\label{tab:verdict}
\end{table}

\paragraph{Training on our datasets produces the best open Armenian
model.} \ourmodel{} reaches a suite mean of 0.50, above the unadapted
Gemma-4-E4B at 0.48 and above every existing open Armenian model, whose
means range from 0.35 to 0.47 (Figure~\ref{fig:headline},
Table~\ref{tab:baselines}). To our knowledge it is the only open
Armenian-adapted model that ends above its own base, and the only one
trained with verified translated STEM data. HellaSwag-hy and ARC-hy sit at the four-way chance floor for every
model and MultiBLiMP is near ceiling, so the discriminative signal
concentrates in the three knowledge tasks, where \ourmodel{} posts the
best score of any open Armenian model on each and the margin over the
base widens (0.459 to 0.503 on the m-MMLU/INCLUDE/Belebele submean). The
rest of this section traces that margin.

\begin{figure}[t]
\centering
\includegraphics[width=0.78\columnwidth]{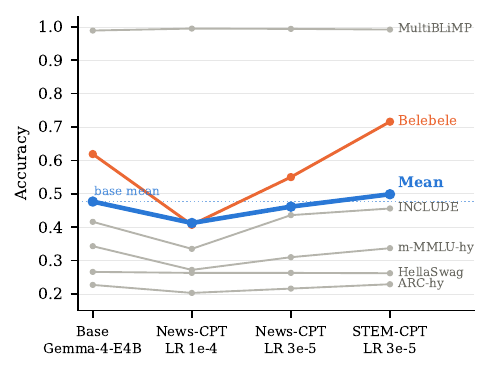}
\caption{Mean accuracy over the six-task suite across CPT runs. Each
point is an independent run and connecting lines are visual guides. The
remaining runs appear in Table~\ref{tab:verdict}.}
\label{fig:forgetting}
\end{figure}

\paragraph{Naive adaptation forgets, early and fast.}
Table~\ref{tab:verdict} and Figure~\ref{fig:forgetting} trace the arc.
\newsrun{} at learning rate $10^{-4}$ costs 21.2 points on Belebele and
7.1 on m-MMLU-hy. In our checkpoint inspections the degradation was
already complete by mid-training, so checkpoint selection cannot recover
it. The released training logs contain the full curves.

\paragraph{A gentler learning rate buys back most, not all.} At
$3{\times}10^{-5}$ the model keeps the full fluency gain and recovers
two-thirds of the Belebele loss. It even surpasses the base on INCLUDE,
whose Armenian-context questions benefit from news knowledge.

\paragraph{Verified translated STEM data reverses the forgetting.}
Swapping 6\% of the news share for \armstem{} lifts the mean 2.2 points
\emph{above} the unadapted base. Belebele reaches 0.716, a gain of 9.7
points over the base, INCLUDE gains 4.0 points, and
m-MMLU-hy returns to within its confidence interval of the base. The
STEM swap also works at the aggressive learning rate, where \stemrun{} at
$10^{-4}$ recovers 18.3 points of Belebele over \newsrun{} at $10^{-4}$
and lifts the mean from 0.412 to 0.462, although in our single-run design
ending \emph{above} the base also required the gentler schedule. The
effect is also saturated at this dose. \fullrun, which draws its STEM
share from the full 373K-item corpus and thereby cuts repetition from
7--9 reads per token to about 1.3 for Armenian and 1.6 for English,
matches \ourmodel{} within confidence
intervals (mean 0.494 against 0.500), so repeating the verified STEM
pool 7--9 times costs nothing on the likelihood suite. On generative
ArmBench the comparison splits. \fullrun{} uniquely lifts exam
mathematics from 1.75 to 2.75 points yet trails on the format-sensitive
tasks (0.57 against 0.62 when its scores in Table~\ref{tab:armbench-full}
are averaged over the twelve 0--1 tasks of Table~\ref{tab:armbencheval}) and costs exam history (2.50 to 0.75).
This suggests that repeated exposure to a verified QA pool doubles as
format training while the full corpus's diversity helps content-heavy
mathematics, with the caveat that \fullrun{} also shifts the STEM
stream's math-to-science ratio (51/49 by items in the training subset
against 15/85 in the full corpus), so composition and repetition change
together. The effect is attributable to the swap itself because the 20\%
English replay stream, the conventional forgetting mitigation
\citep{rolnick2019replay,ibrahim2024cpt}, was present in every run and
did not prevent forgetting. Nor is the reversal a translationese
artifact \citep{globalmmlu2024}, since the gains extend to natively
authored evaluations such as INCLUDE and ArmBench's SynDARin, Hartak, and
national exams (Table~\ref{tab:armbencheval}).

\begin{figure}[t]
\centering
\includegraphics[width=0.80\columnwidth]{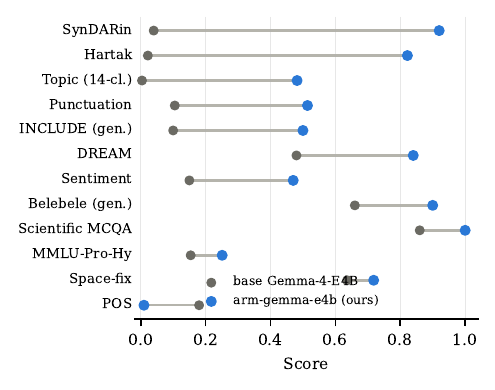}
\caption{Base model vs.\ \ourmodel{} on ArmBench accuracy-style tasks,
sorted by improvement.}
\label{fig:armbench}
\end{figure}

\paragraph{The gains transfer to generative evaluation.} On
capability-measurable ArmBench tasks (Table~\ref{tab:armbencheval},
Figure~\ref{fig:armbench}), \ourmodel{} improves over its base nearly
everywhere, often dramatically. SynDARin rises from 0.04 to 0.92, Belebele
from 0.66 to 0.90, MMLU-Pro-Hy from 0.154 to 0.251, and national-exam
history from 1.0 to 2.5 points. The cross-model comparison makes the
mechanism visible. The base-style competitors collapse on these
strict-format tasks, with tweety-7b at zero almost everywhere and
HyGPT-10b close behind despite its ten billion CPT tokens, because base
models cannot express what they know in the required formats.
ArmenianGPT-1.0-3B, the one instruction-tuned model in the table, is
competitive, yet \ourmodel, without any instruction tuning, posts the
better accuracy-task mean (0.62 against 0.57) while carrying more knowledge on the likelihood
suite (Table~\ref{tab:baselines}); given the small per-task sizes we read
the aggregate rather than single rows. QA-shaped training data teaches
answer discipline without any instruction tuning. The clear competitor
wins are exam literature, where HyGPT leads, plausibly reflecting its
undisclosed corpus, and POS tagging, where the unadapted base is best and
continued pretraining hurts every adapted model.

\paragraph{Existing Armenian models show the same trade at scale.}
tweety-7b, built by trans-tokenization, is fluent at 0.934 MultiBLiMP
yet near chance on every knowledge task. HyGPT-10b, a Gemma-2-9B
continually pretrained on roughly 10B undisclosed Armenian tokens
\citep{hygpt2025}, gains fluency over its own base (MultiBLiMP
$+2.5$pp, outside its $\pm$0.5pp interval) while losing 17.1 points on
Belebele and 4.6 on the suite mean
(Table~\ref{tab:baselines}). Its recipe is undisclosed, so we cannot
isolate the cause, but the pattern of rising fluency and falling knowledge
is exactly the signature above, and the unauditability is itself part of
our argument for open recipes. ArmenianGPT-1.0-3B, the strongest existing open model at a mean of
0.47, still trails even the unadapted Gemma-4-E4B at 0.48. Openly
validated data, not scale, separates \ourmodel{} from this field.

\begin{table}[t]\centering\small
\setlength{\tabcolsep}{4pt}
\begin{tabular}{lrrrrr}
\toprule
 & \multicolumn{2}{c}{HyGPT pair} & & & \\
\cmidrule(lr){2-3}
Task & {\scriptsize G2-9B} & {\scriptsize HyGPT} & {\scriptsize tweety} & {\scriptsize ArmGPT} & {\scriptsize \textbf{ours}} \\
\midrule
MultiBLiMP & 0.971 & \textbf{0.996} & 0.934 & 0.965 & 0.992 \\
HellaSwag-hy & \textbf{0.265} & 0.262 & 0.249 & 0.260 & 0.262 \\
ARC-hy & \textbf{0.231} & 0.219 & 0.217 & 0.205 & 0.229 \\
m-MMLU-hy & 0.329 & 0.282 & 0.226 & 0.330 & \textbf{0.337} \\
INCLUDE & 0.433 & 0.367 & 0.262 & 0.413 & \textbf{0.456} \\
Belebele & 0.660 & 0.489 & 0.229 & 0.654 & \textbf{0.716} \\
\midrule
Mean & 0.48 & 0.44 & 0.35 & 0.47 & \textbf{0.50} \\
\bottomrule
\end{tabular}
\caption{Open Armenian models on the likelihood suite (accuracy, row
best in bold). G2-9B is Gemma-2-9B, HyGPT-10b's own base model;
\textbf{ours} is \ourmodel.}
\label{tab:baselines}
\end{table}

\begin{table}[t]\centering\small
\setlength{\tabcolsep}{2.6pt}
\begin{tabular}{lrrrrr}
\toprule
ArmBench task & {\scriptsize tweety} & {\scriptsize HyGPT} & {\scriptsize ArmGPT$^\dagger$} & {\scriptsize base} & {\scriptsize \textbf{ours}} \\
\midrule
Scientific MCQA & 0.000 & 0.300 & \textbf{1.000} & 0.860 & \textbf{1.000}$^\ast$ \\
Belebele (gen.) & 0.000 & 0.200 & 0.800 & 0.660 & \textbf{0.900} \\
SynDARin & 0.000 & 0.340 & \textbf{0.920} & 0.040 & \textbf{0.920} \\
DREAM & 0.000 & 0.220 & 0.700 & 0.480 & \textbf{0.840} \\
Hartak & 0.000 & 0.022 & \textbf{0.822} & 0.022 & \textbf{0.822} \\
MMLU-Pro-Hy & 0.000 & 0.026 & \textbf{0.281} & 0.154 & 0.251 \\
Exam history & 0.50 & \textbf{2.50} & 2.00 & 1.00 & \textbf{2.50} \\
Exam literature & 0.50 & \textbf{4.25} & 3.00 & 3.00 & 3.25 \\
Topic (14-class) & 0.000 & 0.071 & \textbf{0.504} & 0.004 & 0.482 \\
Punctuation & 0.000 & 0.000 & 0.325 & 0.105 & \textbf{0.514} \\
INCLUDE (gen.) & 0.000 & 0.060 & 0.440 & 0.100 & \textbf{0.500} \\
Sentiment & 0.000 & 0.250 & \textbf{0.550} & 0.150 & 0.470 \\
Space-fix & 0.042 & 0.419 & 0.535 & 0.636 & \textbf{0.718} \\
POS & 0.000 & 0.000 & 0.010 & \textbf{0.180} & 0.010 \\
\midrule
Mean (0--1 tasks) & 0.00 & 0.16 & 0.57 & 0.28 & \textbf{0.62} \\
\bottomrule
\end{tabular}
\caption{ArmBench tasks with accuracy-style metrics (row best in bold,
ties both). BLEU- and judge-scored tasks appear in
Appendix~\ref{app:armbench}. Task sizes: 45--50 items for MCQA rows, 100--280 for
Sentiment/Punctuation/Topic, $n{\approx}1{,}000$ for MMLU-Pro-Hy.
$^\dagger$Instruction-tuned ArmenianGPT-1.0-3B, evaluated as described in
Appendix~\ref{app:repro}, without a chat template like all models here. $^\ast$Audited:
zero shared 8-grams with \armstem{} (Appendix~\ref{app:audit}).}
\label{tab:armbencheval}
\end{table}

Negative results, including a regression on POS tagging and flat exam
mathematics, are reported in Appendix~\ref{app:negatives}.

\section{Related work}
\label{sec:related}

\paragraph{Language-specific ecosystems.} Massively multilingual models
\citep{bloom2022,aya2024} and instruction collections \citep{ayadataset2024}
concentrate capacity in the high-resource head, and small monolingual
models can beat far larger multilingual ones \citep{goldfish2026}. The corpus+model+evaluation release pattern we follow was established by
Latxa \citep{latxa2024} and extended to Kazakh \citep{sherkala2025},
Southeast Asian languages \citep{seallms2024}, Finnish \citep{fingpt2023},
Estonian \citep{llammas2024}, Japanese \citep{swallow2024}, and Polish
\citep{pllum2025}, a design space that \citet{tejaswi2024} systematize.
For Armenian, prior work releases models without their training data,
with HyGPT-10b's $\sim$10B-token corpus undisclosed \citep{hygpt2025} and
tweety-7b releasing its trans-tokenization recipe but not its Armenian
continued-pretraining corpus \citep{remy2024transtok}, while corpus
resources are the
non-redistributable EANC \citep{eanc2022} and Armenian slices of
multilingual crawls
\citep{oscar2022,penedo2025fineweb2,hplt2025,madlad2023,mt5_2021,
cc100_2020,glot500_2023}, of which our audit covers the three largest
recent ones. Evaluation infrastructure has recently arrived
\citep{armbench2025,syndarin2025,belebele2024,include2025,
adelani2024sib200,multiblimp2026}, making the missing piece the open
training data this paper supplies.

\paragraph{Corpus curation and hygiene.} Web-scale pipelines evolved
from heuristic filtering \citep{raffel2020c4,rae2021gopher} to
dedup-centric \citep{refinedweb2023} and empirically ablated
\citep{dclm2024,penedo2025fineweb2} recipes, with documentation as an
artifact \citep{gebru2021datasheets,dolma2024,hplt2025}. Cross-source global deduplication
\citep{slimpajama2023} and syndication-aware thresholds for news
\citep{silcock2023newsdedup} are the two refinements our pipeline builds on.
Contamination auditing \citep{dodge2021c4audit,sainz2023contamination,
oren2024contamination} motivates our decontamination gates and the scan in
Table~\ref{tab:contamination}. Our epoch caps follow data-constrained
scaling laws \citep{muennighoff2023dataconstrained,hernandez2022repetition}.

\paragraph{Continued pretraining and forgetting.} Domain- and
language-adaptive pretraining
\citep{gururangan2020dapt,cui2023chinesellama,sambalingo2024} inherits
catastrophic forgetting
\citep{mccloskey1989forgetting,french1999forgetting,kirkpatrick2017ewc},
conventionally mitigated by replay \citep{rolnick2019replay} and
learning-rate choices
\citep{gupta2023rewarm,ibrahim2024cpt,parmar2024reuse}. Tokenizer
adaptation is the other axis, with dedicated embedding initializers
\citep{wechsel2022,focus2023} and trans-tokenization
\citep{remy2024transtok} as alternatives to the mean initialization we
ablate (\S\ref{sec:model}). We contribute a third mitigation lever,
verified translated knowledge data in the mixture, which does not merely
slow forgetting like replay but reverses it.

\paragraph{Translated training data.} Chain-of-thought transfers across
languages \citep{mgsm2023}, translated math data improves multilingual
reasoning \citep{mathoctopus2024}, translated traces can beat native
distillation \citep{barua2025translatedcot}, multilingual instruction data
is built largely by translation \citep{bactrianx2023,aya2024}, and
parallel data also helps at the pretraining stage \citep{transwebedu2024}.
Prior work verifies translations by LLM judging \citep{gemba2023} or
formula integrity \citep{mathoctopus2024} and answer-checks
\emph{generated} math data \citep{openmathinstruct2_2024}. Our blind
re-solve extends answer-checking to translation, and we deploy translated
data in \emph{continued pretraining} rather than instruction tuning.

\section{Conclusion}
We release two rigorously validated Armenian datasets and show that
training on them produces the strongest open Armenian model, the first
published with its complete corpus and recipe, and that verified
translated STEM data turns the fluency-for-knowledge trade of low-resource
adaptation into a net gain. The pipeline uses no Armenian-specific
machinery and should transfer to other low-resource languages. Since every
artifact carries per-item provenance and licenses, the ecosystem can grow
in place, and a competition-math tranche of \armstem{} and an
instruction-tuned \ourmodel{} are the next releases.

\section*{Limitations}
\label{sec:limitations}
\armweb{} is news-domain-concentrated, so register diversity is limited.
At 3.3B Gemma tokens it is about a third of the undisclosed
$\sim$10B-token corpus behind HyGPT, if that count is in Gemma-2 tokens, a
comparable unit. We argue curation and openness over scale but note the gap.
Within the 6-point mixture swap we cannot yet separate the contributions
of STEM content, QA format, and verification, and a format-matched
comparison run is future work (\fullrun{} already provides a
repetition-matched one, \S\ref{sec:results}). Evaluation is MCQA-heavy with no generation quality or
safety assessment. The perfect scientific-MCQA score passes n-gram
decontamination and a direct per-item n-gram audit against \armstem{}
(Appendix~\ref{app:audit}), but the 50-item set is small and a
paraphrase-level audit is future work.

\section*{Ethics Statement}
\textbf{Copyright and licensing.} \armweb{} redistributes the text of
publicly published news articles crawled by one of the authors over 15 years. The
\emph{compilation} (selection, cleaning, deduplication, metadata, splits) is
released under ODC-BY~1.0, while copyright in the underlying articles
remains with their publishers, and we make no claim over it. The crawl honored the
sites' crawling rules, including robots.txt directives. The release is made
for research use in the text-and-data-mining tradition under which C4,
OSCAR, HPLT, and FineWeb-class corpora are distributed
\citep{raffel2020c4,oscar2022,hplt2025}. Per-document \texttt{url} and
\texttt{source} fields preserve attribution, and the hosting repository
provides a takedown mechanism through which any rights holder can have
their content removed. \textbf{Personal data.} Author bylines are removed
from all records. A scan of the released text finds contact-style PII in a
small fraction of documents, with 942 e-mail addresses in 782 documents
and 8{,}971 documents containing phone-number-like strings, overwhelmingly
newsroom and institutional contact lines that are part of the published
articles. The scan report ships with the corpus. News text inherently
discusses named individuals, and we release only what the source outlets
made public.
\textbf{\armstem} inherits its sources' MIT/CC-BY-4.0 licenses with a
statement of changes (machine translation with automated verification), and
its card notes that OpenScience/OSR-2 are Qwen-generated synthetic data.
\textbf{Intended use.} The corpora and model are research artifacts. The
model is a base model with no safety tuning and should not be deployed to
end users without further alignment.

\bibliography{custom}

\begin{thebibliography}{94}
\providecommand{\natexlab}[1]{#1}

\bibitem[{Abadji et~al.(2022)Abadji, Ortiz~Suarez, Romary, and
  Sagot}]{oscar2022}
Julien Abadji, Pedro Ortiz~Suarez, Laurent Romary, and Beno{\^i}t Sagot. 2022.
\newblock Towards a cleaner document-oriented multilingual crawled corpus.
\newblock In \emph{Proceedings of the 13th Language Resources and Evaluation
  Conference (LREC)}.
\newblock ArXiv:2201.06642.

\bibitem[{Adelani et~al.(2024)Adelani, Liu, Shen, Vassilyev, Alabi, Mao, Gao,
  and Lee}]{adelani2024sib200}
David~Ifeoluwa Adelani, Hannah Liu, Xiaoyu Shen, Nikita Vassilyev, Jesujoba~O.
  Alabi, Yanke Mao, Haonan Gao, and En-Shiun~Annie Lee. 2024.
\newblock \href {https://doi.org/10.18653/v1/2024.eacl-long.14} {{SIB}-200: A
  simple, inclusive, and big evaluation dataset for topic classification in
  200+ languages and dialects}.
\newblock In \emph{Proceedings of the 18th Conference of the European Chapter
  of the Association for Computational Linguistics (Volume 1: Long Papers)},
  pages 226--245.

\bibitem[{Ahia et~al.(2023)Ahia, Kumar, Gonen, Kasai, Mortensen, Smith, and
  Tsvetkov}]{ahia2023tokencost}
Orevaoghene Ahia, Sachin Kumar, Hila Gonen, Jungo Kasai, David~R. Mortensen,
  Noah~A. Smith, and Yulia Tsvetkov. 2023.
\newblock Do all languages cost the same? tokenization in the era of commercial
  language models.
\newblock In \emph{Proceedings of the 2023 Conference on Empirical Methods in
  Natural Language Processing}, pages 9904--9923.
\newblock ArXiv:2305.13707.

\bibitem[{{ArmGPT}(2026)}]{armeniangpt2026}
{ArmGPT}. 2026.
\newblock Armeniangpt-1.0-3b.
\newblock Hugging Face model card,
  \url{https://huggingface.co/ArmGPT/ArmenianGPT-1.0-3B}.

\bibitem[{Bandarkar et~al.(2024)Bandarkar, Liang, Muller, Artetxe, Shukla,
  Husa, Goyal, Krishnan, Zettlemoyer, and Khabsa}]{belebele2024}
Lucas Bandarkar, Davis Liang, Benjamin Muller, Mikel Artetxe, Satya~Narayan
  Shukla, Donald Husa, Naman Goyal, Abhinandan Krishnan, Luke Zettlemoyer, and
  Madian Khabsa. 2024.
\newblock \href {https://doi.org/10.18653/v1/2024.acl-long.44} {The belebele
  benchmark: a parallel reading comprehension dataset in 122 language
  variants}.
\newblock In \emph{Proceedings of the 62nd Annual Meeting of the Association
  for Computational Linguistics (Volume 1: Long Papers)}, pages 749--775.

\bibitem[{Barua et~al.(2025)Barua, Eisape, Yin, and
  Suhr}]{barua2025translatedcot}
Josh Barua, Seun Eisape, Kayo Yin, and Alane Suhr. 2025.
\newblock Long chain-of-thought reasoning across languages.
\newblock \emph{arXiv preprint arXiv:2508.14828}.

\bibitem[{Burchell et~al.(2025)Burchell, de~Gibert, Arefyev, Aulamo,
  Ba{\~n}{\'o}n, Chen, Fedorova, Guillou, Haddow, Haji{\v c} et~al.}]{hplt2025}
Laurie Burchell, Ona de~Gibert, Nikolay Arefyev, Mikko Aulamo, Marta
  Ba{\~n}{\'o}n, Pinzhen Chen, Mariia Fedorova, Liane Guillou, Barry Haddow,
  Jan Haji{\v c}, et~al. 2025.
\newblock An expanded massive multilingual dataset for high-performance
  language technologies ({HPLT}).
\newblock In \emph{Proceedings of ACL 2025}.
\newblock ArXiv:2503.10267; anthology 2025.acl-long.854.

\bibitem[{Chang et~al.(2026)Chang, Arnett, Tu, and Bergen}]{goldfish2026}
Tyler~A. Chang, Catherine Arnett, Zhuowen Tu, and Benjamin~K. Bergen. 2026.
\newblock Goldfish: Monolingual language models for 350 languages.
\newblock In \emph{Proceedings of the Language Resources and Evaluation
  Conference (LREC 2026)}.
\newblock ArXiv:2408.10441.

\bibitem[{Chen et~al.(2024)Chen, Zheng, Wu, Gong, Zhang, and
  Li}]{mathoctopus2024}
Nuo Chen, Zinan Zheng, Ning Wu, Ming Gong, Dongmei Zhang, and Jia Li. 2024.
\newblock \href {https://aclanthology.org/2024.findings-emnlp.411/} {Breaking
  language barriers in multilingual mathematical reasoning: Insights and
  observations}.
\newblock In \emph{Findings of the Association for Computational Linguistics:
  EMNLP 2024}.
\newblock ArXiv:2310.20246.

\bibitem[{Chen et~al.(2025)Chen, Prabhumoye, Bercovich et~al.}]{acereason2025}
Yang Chen, Shrimai Prabhumoye, Akhiad Bercovich, et~al. 2025.
\newblock Acereason-nemotron: Advancing math and code reasoning through
  reinforcement learning.
\newblock \emph{arXiv preprint arXiv:2505.16400}.

\bibitem[{Clark et~al.(2018)Clark, Cowhey, Etzioni, Khot, Sabharwal, Schoenick,
  and Tafjord}]{arc2018}
Peter Clark, Isaac Cowhey, Oren Etzioni, Tushar Khot, Ashish Sabharwal, Carissa
  Schoenick, and Oyvind Tafjord. 2018.
\newblock \href {https://arxiv.org/abs/1803.05457} {Think you have solved
  question answering? try {ARC}, the {AI2} reasoning challenge}.
\newblock \emph{Preprint}, arXiv:1803.05457.

\bibitem[{Cobbe et~al.(2021)Cobbe, Kosaraju, Bavarian, Chen, Jun, Kaiser,
  Plappert, Tworek, Hilton, Nakano, Hesse, and Schulman}]{gsm8k2021}
Karl Cobbe, Vineet Kosaraju, Mohammad Bavarian, Mark Chen, Heewoo Jun, Lukasz
  Kaiser, Matthias Plappert, Jerry Tworek, Jacob Hilton, Reiichiro Nakano,
  Christopher Hesse, and John Schulman. 2021.
\newblock \href {https://arxiv.org/abs/2110.14168} {Training verifiers to solve
  math word problems}.
\newblock \emph{Preprint}, arXiv:2110.14168.

\bibitem[{Conneau et~al.(2020)Conneau, Khandelwal, Goyal, Chaudhary, Wenzek,
  Guzm{\'a}n, Grave, Ott, Zettlemoyer, and Stoyanov}]{cc100_2020}
Alexis Conneau, Kartikay Khandelwal, Naman Goyal, Vishrav Chaudhary, Guillaume
  Wenzek, Francisco Guzm{\'a}n, Edouard Grave, Myle Ott, Luke Zettlemoyer, and
  Veselin Stoyanov. 2020.
\newblock Unsupervised cross-lingual representation learning at scale.
\newblock In \emph{Proceedings of the 58th Annual Meeting of the Association
  for Computational Linguistics (ACL)}.
\newblock ArXiv:1911.02116.

\bibitem[{Crego et~al.(2016)Crego, Kim, Klein, Rebollo, Yang, Senellart
  et~al.}]{crego2016systran}
Josep Crego, Jungi Kim, Guillaume Klein, Anabel Rebollo, Kathy Yang, Jean
  Senellart, et~al. 2016.
\newblock Systran's pure neural machine translation systems.
\newblock \emph{arXiv preprint arXiv:1610.05540}.

\bibitem[{Csaki et~al.(2024)Csaki, Li, Li, Xu, Pawakapan, Zhang, Du, Zhao, Hu,
  and Thakker}]{sambalingo2024}
Zoltan Csaki, Bo~Li, Jonathan Li, Qiantong Xu, Pian Pawakapan, Leon Zhang, Yun
  Du, Hengyu Zhao, Changran Hu, and Urmish Thakker. 2024.
\newblock Sambalingo: Teaching large language models new languages.
\newblock \emph{arXiv preprint arXiv:2404.05829}.

\bibitem[{Cui et~al.(2023)Cui, Yang, and Yao}]{cui2023chinesellama}
Yiming Cui, Ziqing Yang, and Xin Yao. 2023.
\newblock Efficient and effective text encoding for chinese llama and alpaca.
\newblock \emph{arXiv preprint arXiv:2304.08177}.

\bibitem[{Dinu et~al.(2019)Dinu, Mathur, Federico, and
  Al-Onaizan}]{dinu2019terminology}
Georgiana Dinu, Prashant Mathur, Marcello Federico, and Yaser Al-Onaizan. 2019.
\newblock Training neural machine translation to apply terminology constraints.
\newblock In \emph{Proceedings of the 57th Annual Meeting of the Association
  for Computational Linguistics (ACL)}.
\newblock ArXiv:1906.01105.

\bibitem[{Dobler and de~Melo(2023)}]{focus2023}
Konstantin Dobler and Gerard de~Melo. 2023.
\newblock Focus: Effective embedding initialization for monolingual
  specialization of multilingual models.
\newblock In \emph{Proceedings of the 2023 Conference on Empirical Methods in
  Natural Language Processing (EMNLP)}.
\newblock ArXiv:2305.14481.

\bibitem[{Dodge et~al.(2021)Dodge, Sap, Marasovi{\'c}, Agnew, Ilharco,
  Groeneveld, Mitchell, and Gardner}]{dodge2021c4audit}
Jesse Dodge, Maarten Sap, Ana Marasovi{\'c}, William Agnew, Gabriel Ilharco,
  Dirk Groeneveld, Margaret Mitchell, and Matt Gardner. 2021.
\newblock Documenting large webtext corpora: A case study on the colossal clean
  crawled corpus.
\newblock In \emph{Proceedings of the 2021 Conference on Empirical Methods in
  Natural Language Processing (EMNLP)}.
\newblock ArXiv:2104.08758.

\bibitem[{Etxaniz et~al.(2024)Etxaniz, Sainz, Perez, Aldabe, Rigau, Agirre,
  Ormazabal, Artetxe, and Soroa}]{latxa2024}
Julen Etxaniz, Oscar Sainz, Naiara Perez, Itziar Aldabe, German Rigau, Eneko
  Agirre, Aitor Ormazabal, Mikel Artetxe, and Aitor Soroa. 2024.
\newblock Latxa: An open language model and evaluation suite for basque.
\newblock In \emph{Proceedings of the 62nd Annual Meeting of the Association
  for Computational Linguistics (Volume 1: Long Papers)}, pages 14952--14972.
\newblock ArXiv:2403.20266.

\bibitem[{French(1999)}]{french1999forgetting}
Robert~M. French. 1999.
\newblock Catastrophic forgetting in connectionist networks.
\newblock \emph{Trends in Cognitive Sciences}, 3(4):128--135.

\bibitem[{Fujii et~al.(2024)Fujii, Nakamura, Loem, Iida, Ohi, Hattori, Shota,
  Mizuki, Yokota, and Okazaki}]{swallow2024}
Kazuki Fujii, Taishi Nakamura, Mengsay Loem, Hiroki Iida, Masanari Ohi, Kakeru
  Hattori, Hirai Shota, Sakae Mizuki, Rio Yokota, and Naoaki Okazaki. 2024.
\newblock Continual pre-training for cross-lingual llm adaptation: Enhancing
  japanese language capabilities.
\newblock In \emph{Conference on Language Modeling (COLM)}.
\newblock ArXiv:2404.17790.

\bibitem[{Gao et~al.(2024)Gao, Tow, Abbasi, Biderman, Black, DiPofi, Foster,
  Golding, Hsu, Le~Noac'h, Li, McDonell, Muennighoff, Ociepa, Phang, Reynolds,
  Schoelkopf, Skowron, Sutawika, Tang, Thite, Wang, Wang, and
  Zou}]{evalharness2024}
Leo Gao, Jonathan Tow, Baber Abbasi, Stella Biderman, Sid Black, Anthony
  DiPofi, Charles Foster, Laurence Golding, Jeffrey Hsu, Alain Le~Noac'h,
  Haonan Li, Kyle McDonell, Niklas Muennighoff, Chris Ociepa, Jason Phang,
  Laria Reynolds, Hailey Schoelkopf, Aviya Skowron, Lintang Sutawika, and 5
  others. 2024.
\newblock \href {https://doi.org/10.5281/zenodo.12608602} {The language model
  evaluation harness}.

\bibitem[{Gebru et~al.(2021)Gebru, Morgenstern, Vecchione, Vaughan, Wallach,
  Daum{\'e}~III, and Crawford}]{gebru2021datasheets}
Timnit Gebru, Jamie Morgenstern, Briana Vecchione, Jennifer~Wortman Vaughan,
  Hanna Wallach, Hal Daum{\'e}~III, and Kate Crawford. 2021.
\newblock Datasheets for datasets.
\newblock \emph{Communications of the ACM}, 64(12):86--92.
\newblock ArXiv:1803.09010.

\bibitem[{{Gen2B} and {NCCAIT}(2025)}]{hygpt2025}
{Gen2B} and {NCCAIT}. 2025.
\newblock Hygpt-10b: A large language model for eastern armenian.
\newblock Hugging Face model card,
  \url{https://huggingface.co/Gen2B/HyGPT-10b}.
\newblock Technical report at gen2b.ai/hygpt-release-1-0.

\bibitem[{Ghazaryan et~al.(2025)Ghazaryan, Arakelyan, Augenstein, and
  Minervini}]{syndarin2025}
Gayane Ghazaryan, Erik Arakelyan, Isabelle Augenstein, and Pasquale Minervini.
  2025.
\newblock \href {https://aclanthology.org/2025.coling-main.430/} {{S}yn{DAR}in:
  Synthesising datasets for automated reasoning in low-resource languages}.
\newblock In \emph{Proceedings of the 31st International Conference on
  Computational Linguistics (COLING)}, pages 6459--6466.
\newblock ArXiv:2406.14425.

\bibitem[{{Google DeepMind}(2026)}]{gemini31_2026}
{Google DeepMind}. 2026.
\newblock \href
  {https://deepmind.google/models/model-cards/gemini-3-1-flash-lite/} {{Gemini
  3.1 Flash-Lite} model card}.

\bibitem[{Gupta et~al.(2023)Gupta, Th{\'e}rien, Ibrahim, Richter, Anthony,
  Belilovsky, Rish, and Lesort}]{gupta2023rewarm}
Kshitij Gupta, Benjamin Th{\'e}rien, Adam Ibrahim, Mats~L. Richter, Quentin
  Anthony, Eugene Belilovsky, Irina Rish, and Timoth{\'e}e Lesort. 2023.
\newblock Continual pre-training of large language models: How to (re)warm your
  model?
\newblock \emph{arXiv preprint arXiv:2308.04014}.

\bibitem[{Gururangan et~al.(2020)Gururangan, Marasovi{\'c}, Swayamdipta, Lo,
  Beltagy, Downey, and Smith}]{gururangan2020dapt}
Suchin Gururangan, Ana Marasovi{\'c}, Swabha Swayamdipta, Kyle Lo, Iz~Beltagy,
  Doug Downey, and Noah~A. Smith. 2020.
\newblock Don't stop pretraining: Adapt language models to domains and tasks.
\newblock In \emph{Proceedings of the 58th Annual Meeting of the Association
  for Computational Linguistics}, pages 8342--8360.

\bibitem[{Habib et~al.(2023)Habib, Fourrier, Kydl{\'i}{\v{c}}ek, Wolf, and
  Tunstall}]{lighteval2023}
Nathan Habib, Cl{\'e}mentine Fourrier, Hynek Kydl{\'i}{\v{c}}ek, Thomas Wolf,
  and Lewis Tunstall. 2023.
\newblock \href {https://github.com/huggingface/lighteval} {Lighteval: A
  lightweight framework for llm evaluation}.

\bibitem[{Hendrycks et~al.(2021)Hendrycks, Burns, Basart, Zou, Mazeika, Song,
  and Steinhardt}]{mmlu2021}
Dan Hendrycks, Collin Burns, Steven Basart, Andy Zou, Mantas Mazeika, Dawn
  Song, and Jacob Steinhardt. 2021.
\newblock Measuring massive multitask language understanding.
\newblock In \emph{International Conference on Learning Representations
  (ICLR)}.
\newblock ArXiv:2009.03300.

\bibitem[{Hernandez et~al.(2022)Hernandez, Brown, Conerly, DasSarma, Drain,
  El-Showk, Elhage, Hatfield-Dodds, Henighan, Hume
  et~al.}]{hernandez2022repetition}
Danny Hernandez, Tom Brown, Tom Conerly, Nova DasSarma, Dawn Drain, Sheer
  El-Showk, Nelson Elhage, Zac Hatfield-Dodds, Tom Henighan, Tristan Hume,
  et~al. 2022.
\newblock Scaling laws and interpretability of learning from repeated data.
\newblock \emph{arXiv preprint arXiv:2205.10487}.

\bibitem[{Hoffmann et~al.(2022)Hoffmann, Borgeaud, Mensch, Buchatskaya, Cai,
  Rutherford, de~Las~Casas, Hendricks, Welbl, Clark
  et~al.}]{hoffmann2022chinchilla}
Jordan Hoffmann, Sebastian Borgeaud, Arthur Mensch, Elena Buchatskaya, Trevor
  Cai, Eliza Rutherford, Diego de~Las~Casas, Lisa~Anne Hendricks, Johannes
  Welbl, Aidan Clark, et~al. 2022.
\newblock Training compute-optimal large language models.
\newblock \emph{arXiv preprint arXiv:2203.15556}.

\bibitem[{Ibrahim et~al.(2024)Ibrahim, Th{\'e}rien, Gupta, Richter, Anthony,
  Lesort, Belilovsky, and Rish}]{ibrahim2024cpt}
Adam Ibrahim, Benjamin Th{\'e}rien, Kshitij Gupta, Mats~L. Richter, Quentin
  Anthony, Timoth{\'e}e Lesort, Eugene Belilovsky, and Irina Rish. 2024.
\newblock Simple and scalable strategies to continually pre-train large
  language models.
\newblock \emph{Transactions on Machine Learning Research}.
\newblock ArXiv:2403.08763.

\bibitem[{Imani et~al.(2023)Imani, Lin, Kargaran, Severini, Jalili~Sabet,
  Kassner, Ma, Schmid, Martins, Yvon, and Sch{\"u}tze}]{glot500_2023}
Ayyoob Imani, Peiqin Lin, Amir~Hossein Kargaran, Silvia Severini, Masoud
  Jalili~Sabet, Nora Kassner, Chunlan Ma, Helmut Schmid, Andr{\'e} Martins,
  Fran{\c{c}}ois Yvon, and Hinrich Sch{\"u}tze. 2023.
\newblock Glot500: Scaling multilingual corpora and language models to 500
  languages.
\newblock In \emph{Proceedings of the 61st Annual Meeting of the Association
  for Computational Linguistics (ACL)}.
\newblock ArXiv:2305.12182.

\bibitem[{Jumelet et~al.(2026)Jumelet, Weissweiler, Nivre, and
  Bisazza}]{multiblimp2026}
Jaap Jumelet, Leonie Weissweiler, Joakim Nivre, and Arianna Bisazza. 2026.
\newblock \href {https://doi.org/10.1162/tacl.a.600} {{M}ulti{BL}i{MP} 1.0: A
  massively multilingual benchmark of linguistic minimal pairs}.
\newblock \emph{Transactions of the Association for Computational Linguistics},
  14:193--216.
\newblock ArXiv:2504.02768.

\bibitem[{Kandpal et~al.(2022)Kandpal, Wallace, and
  Raffel}]{kandpal2022privacy}
Nikhil Kandpal, Eric Wallace, and Colin Raffel. 2022.
\newblock Deduplicating training data mitigates privacy risks in language
  models.
\newblock In \emph{Proceedings of the 39th International Conference on Machine
  Learning (ICML), PMLR 162}.
\newblock ArXiv:2202.06539.

\bibitem[{Kaplan et~al.(2020)Kaplan, McCandlish, Henighan, Brown, Chess, Child,
  Gray, Radford, Wu, and Amodei}]{kaplan2020scaling}
Jared Kaplan, Sam McCandlish, Tom Henighan, Tom~B. Brown, Benjamin Chess, Rewon
  Child, Scott Gray, Alec Radford, Jeffrey Wu, and Dario Amodei. 2020.
\newblock Scaling laws for neural language models.
\newblock \emph{arXiv preprint arXiv:2001.08361}.

\bibitem[{Kargaran et~al.(2023)Kargaran, Imani, Yvon, and
  Sch{\"u}tze}]{glotlid2023}
Amir~Hossein Kargaran, Ayyoob Imani, Fran{\c c}ois Yvon, and Hinrich
  Sch{\"u}tze. 2023.
\newblock {GlotLID}: Language identification for low-resource languages.
\newblock In \emph{Findings of the Association for Computational Linguistics:
  EMNLP 2023}, pages 6155--6218.
\newblock ArXiv:2310.16248.

\bibitem[{Khurshudyan et~al.(2022)Khurshudyan, Arkhangelskiy, Daniel, Plungian,
  Levonian, Polyakov, and Rubakov}]{eanc2022}
Victoria Khurshudyan, Timofey Arkhangelskiy, Misha Daniel, Vladimir Plungian,
  Dmitri Levonian, Alex Polyakov, and Sergei Rubakov. 2022.
\newblock \href {https://aclanthology.org/2022.digitam-1.5/} {{E}astern
  {A}rmenian national corpus: State of the art and perspectives}.
\newblock In \emph{Proceedings of the Workshop on Processing Language
  Variation: Digital Armenian (DigitAm) at LREC 2022}, pages 28--37.

\bibitem[{Kirkpatrick et~al.(2017)Kirkpatrick, Pascanu, Rabinowitz, Veness,
  Desjardins, Rusu, Milan, Quan, Ramalho, Grabska-Barwinska, Hassabis, Clopath,
  Kumaran, and Hadsell}]{kirkpatrick2017ewc}
James Kirkpatrick, Razvan Pascanu, Neil Rabinowitz, Joel Veness, Guillaume
  Desjardins, Andrei~A. Rusu, Kieran Milan, John Quan, Tiago Ramalho, Agnieszka
  Grabska-Barwinska, Demis Hassabis, Claudia Clopath, Dharshan Kumaran, and
  Raia Hadsell. 2017.
\newblock Overcoming catastrophic forgetting in neural networks.
\newblock \emph{Proceedings of the National Academy of Sciences},
  114(13):3521--3526.

\bibitem[{Kocetkov et~al.(2022)Kocetkov, Li, Ben~Allal, Li, Mou,
  Mu{\~n}oz~Ferrandis, Jernite, Mitchell, Hughes, Wolf, Bahdanau, von Werra,
  and de~Vries}]{thestack2022}
Denis Kocetkov, Raymond Li, Loubna Ben~Allal, Jia Li, Chenghao Mou, Carlos
  Mu{\~n}oz~Ferrandis, Yacine Jernite, Margaret Mitchell, Sean Hughes, Thomas
  Wolf, Dzmitry Bahdanau, Leandro von Werra, and Harm de~Vries. 2022.
\newblock The stack: 3 tb of permissively licensed source code.
\newblock \emph{arXiv preprint arXiv:2211.15533}.

\bibitem[{Kocmi and Federmann(2023)}]{gemba2023}
Tom Kocmi and Christian Federmann. 2023.
\newblock Large language models are state-of-the-art evaluators of translation
  quality.
\newblock In \emph{Proceedings of the 24th Annual Conference of the European
  Association for Machine Translation (EAMT)}.
\newblock ArXiv:2302.14520.

\bibitem[{Koco{\'n} et~al.(2025)Koco{\'n}, Piasecki, Janz, Ferdinan,
  Radli{\'n}ski et~al.}]{pllum2025}
Jan Koco{\'n}, Maciej Piasecki, Arkadiusz Janz, Teddy Ferdinan, {\L}ukasz
  Radli{\'n}ski, et~al. 2025.
\newblock Pllum: A family of polish large language models.
\newblock \emph{arXiv preprint arXiv:2511.03823}.

\bibitem[{Koto et~al.(2025)Koto, Joshi, Mukhituly, Wang et~al.}]{sherkala2025}
Fajri Koto, Rituraj Joshi, Nurdaulet Mukhituly, Yuxia Wang, et~al. 2025.
\newblock Sherkala-chat: Building a state-of-the-art llm for kazakh in a
  moderately resourced setting.
\newblock In \emph{Conference on Language Modeling (COLM)}.
\newblock ArXiv:2503.01493.

\bibitem[{Kudugunta et~al.(2023)Kudugunta, Caswell, Zhang, Garcia,
  Choquette-Choo, Lee, Xin, Kusupati, Stella, Bapna, and Firat}]{madlad2023}
Sneha Kudugunta, Isaac Caswell, Biao Zhang, Xavier Garcia, Christopher~A.
  Choquette-Choo, Katherine Lee, Derrick Xin, Aditya Kusupati, Romi Stella,
  Ankur Bapna, and Orhan Firat. 2023.
\newblock Madlad-400: A multilingual and document-level large audited dataset.
\newblock In \emph{Advances in Neural Information Processing Systems 36
  (Datasets and Benchmarks Track)}.
\newblock ArXiv:2309.04662.

\bibitem[{Kuulmets et~al.(2024)Kuulmets, Purason, Luhtaru, and
  Fishel}]{llammas2024}
Hele-Andra Kuulmets, Taido Purason, Agnes Luhtaru, and Mark Fishel. 2024.
\newblock Teaching llama a new language through cross-lingual knowledge
  transfer.
\newblock In \emph{Findings of the Association for Computational Linguistics:
  NAACL 2024}.
\newblock ArXiv:2404.04042.

\bibitem[{Lai et~al.(2023)Lai, Ngo, Ben~Veyseh, Man, Dernoncourt, Bui, and
  Nguyen}]{okapi2023}
Viet~Dac Lai, Nghia~Trung Ngo, Amir~Pouran Ben~Veyseh, Hieu Man, Franck
  Dernoncourt, Trung Bui, and Thien~Huu Nguyen. 2023.
\newblock Okapi: Instruction-tuned large language models in multiple languages
  with reinforcement learning from human feedback.
\newblock In \emph{Proceedings of the 2023 Conference on Empirical Methods in
  Natural Language Processing: System Demonstrations}.
\newblock ArXiv:2307.16039.

\bibitem[{Le~Scao et~al.(2022)Le~Scao, Fan, Akiki, Pavlick et~al.}]{bloom2022}
Teven Le~Scao, Angela Fan, Christopher Akiki, Ellie Pavlick, et~al. 2022.
\newblock Bloom: A 176b-parameter open-access multilingual language model.
\newblock \emph{arXiv preprint arXiv:2211.05100}.

\bibitem[{Lee et~al.(2022)Lee, Ippolito, Nystrom, Zhang, Eck, Callison-Burch,
  and Carlini}]{lee2022dedup}
Katherine Lee, Daphne Ippolito, Andrew Nystrom, Chiyuan Zhang, Douglas Eck,
  Chris Callison-Burch, and Nicholas Carlini. 2022.
\newblock Deduplicating training data makes language models better.
\newblock In \emph{Proceedings of the 60th Annual Meeting of the Association
  for Computational Linguistics (ACL)}.
\newblock ArXiv:2107.06499.

\bibitem[{Li et~al.(2023)Li, Koto, Wu, Aji, and Baldwin}]{bactrianx2023}
Haonan Li, Fajri Koto, Minghao Wu, Alham~Fikri Aji, and Timothy Baldwin. 2023.
\newblock Bactrian-x: Multilingual replicable instruction-following models with
  low-rank adaptation.
\newblock \emph{arXiv preprint arXiv:2305.15011}.

\bibitem[{Li et~al.(2024)Li, Fang, Smyrnis, Ivgi, Jordan, Gadre, Bansal, Guha,
  Keh, Arora et~al.}]{dclm2024}
Jeffrey Li, Alex Fang, Georgios Smyrnis, Maor Ivgi, Matt Jordan, Samir Gadre,
  Hritik Bansal, Etash Guha, Sedrick Keh, Kushal Arora, et~al. 2024.
\newblock Datacomp-lm: In search of the next generation of training sets for
  language models.
\newblock In \emph{Advances in Neural Information Processing Systems 37
  (Datasets and Benchmarks Track)}.
\newblock ArXiv:2406.11794.

\bibitem[{Luukkonen et~al.(2023)Luukkonen, Komulainen, Luoma, Eskelinen,
  Kanerva, Kupari, Ginter, Laippala, Muennighoff, Piktus et~al.}]{fingpt2023}
Risto Luukkonen, Ville Komulainen, Jouni Luoma, Anni Eskelinen, Jenna Kanerva,
  Hanna-Mari Kupari, Filip Ginter, Veronika Laippala, Niklas Muennighoff,
  Aleksandra Piktus, et~al. 2023.
\newblock Fingpt: Large generative models for a small language.
\newblock In \emph{Proceedings of the 2023 Conference on Empirical Methods in
  Natural Language Processing (EMNLP)}.
\newblock ArXiv:2311.05640.

\bibitem[{Magnusson et~al.(2025)Magnusson, Tai, Bogin, Heineman, Hwang,
  Soldaini, Bhagia, Liu, Groeneveld, Tafjord, Smith, Koh, and
  Dodge}]{datadecide2025}
Ian Magnusson, Nguyen Tai, Ben Bogin, David Heineman, Jena~D. Hwang, Luca
  Soldaini, Akshita Bhagia, Jiacheng Liu, Dirk Groeneveld, Oyvind Tafjord,
  Noah~A. Smith, Pang~Wei Koh, and Jesse Dodge. 2025.
\newblock Datadecide: How to predict best pretraining data with small
  experiments.
\newblock In \emph{Proceedings of the 42nd International Conference on Machine
  Learning (ICML)}.
\newblock ArXiv:2504.11393.

\bibitem[{McCloskey and Cohen(1989)}]{mccloskey1989forgetting}
Michael McCloskey and Neal~J. Cohen. 1989.
\newblock Catastrophic interference in connectionist networks: The sequential
  learning problem.
\newblock In \emph{Psychology of Learning and Motivation}, volume~24, pages
  109--165. Academic Press.

\bibitem[{{Metric AI Lab}(2025)}]{armbench2025}
{Metric AI Lab}. 2025.
\newblock Armbench-llm: Benchmarking llms on armenian language tasks.
\newblock \url{https://huggingface.co/blog/Metric-AI/armbench-llm}.
\newblock Code: \url{https://github.com/Metricam/ArmBench-LLM}.

\bibitem[{Minixhofer et~al.(2022)Minixhofer, Paischer, and
  Rekabsaz}]{wechsel2022}
Benjamin Minixhofer, Fabian Paischer, and Navid Rekabsaz. 2022.
\newblock Wechsel: Effective initialization of subword embeddings for
  cross-lingual transfer of monolingual language models.
\newblock In \emph{Proceedings of the 2022 Conference of the North American
  Chapter of the Association for Computational Linguistics (NAACL)}.
\newblock ArXiv:2112.06598.

\bibitem[{Muennighoff et~al.(2023)Muennighoff, Rush, Barak, Le~Scao, Piktus,
  Tazi, Pyysalo, Wolf, and Raffel}]{muennighoff2023dataconstrained}
Niklas Muennighoff, Alexander~M. Rush, Boaz Barak, Teven Le~Scao, Aleksandra
  Piktus, Nouamane Tazi, Sampo Pyysalo, Thomas Wolf, and Colin Raffel. 2023.
\newblock Scaling data-constrained language models.
\newblock In \emph{Advances in Neural Information Processing Systems 36
  (NeurIPS 2023)}.
\newblock ArXiv:2305.16264.

\bibitem[{Nguyen et~al.(2024{\natexlab{a}})Nguyen, Nguyen, Lai, Man, Ngo,
  Dernoncourt, Rossi, and Nguyen}]{culturax2023}
Thuat Nguyen, Chien~Van Nguyen, Viet~Dac Lai, Hieu Man, Nghia~Trung Ngo, Franck
  Dernoncourt, Ryan~A. Rossi, and Thien~Huu Nguyen. 2024{\natexlab{a}}.
\newblock Culturax: A cleaned, enormous, and multilingual dataset for large
  language models in 167 languages.
\newblock In \emph{Proceedings of the 2024 Joint International Conference on
  Computational Linguistics, Language Resources and Evaluation (LREC-COLING)}.
\newblock ArXiv:2309.09400.

\bibitem[{Nguyen et~al.(2024{\natexlab{b}})Nguyen, Zhang, Li, Aljunied, Hu
  et~al.}]{seallms2024}
Xuan-Phi Nguyen, Wenxuan Zhang, Xin Li, Mahani Aljunied, Zhiqiang Hu, et~al.
  2024{\natexlab{b}}.
\newblock Seallms -- large language models for southeast asia.
\newblock In \emph{Proceedings of the 62nd Annual Meeting of the Association
  for Computational Linguistics (ACL 2024): System Demonstrations}.
\newblock ArXiv:2312.00738.

\bibitem[{{NLLB Team} et~al.(2024){NLLB Team}, Costa-juss{\`a}, Cross,
  {\c{C}}elebi, Elbayad, Heafield, Heffernan, Kalbassi et~al.}]{nllb2024}
{NLLB Team}, Marta~R. Costa-juss{\`a}, James Cross, Onur {\c{C}}elebi, Maha
  Elbayad, Kenneth Heafield, Kevin Heffernan, Elahe Kalbassi, et~al. 2024.
\newblock \href {https://doi.org/10.1038/s41586-024-07335-x} {Scaling neural
  machine translation to 200 languages}.
\newblock \emph{Nature}, 630(8018):841--846.
\newblock ArXiv:2207.04672.

\bibitem[{{NVIDIA}(2025{\natexlab{a}})}]{openscience2025}
{NVIDIA}. 2025{\natexlab{a}}.
\newblock Openscience.
\newblock Hugging Face dataset,
  \url{https://huggingface.co/datasets/nvidia/OpenScience}.

\bibitem[{{NVIDIA}(2025{\natexlab{b}})}]{osr2_2025}
{NVIDIA}. 2025{\natexlab{b}}.
\newblock Opensciencereasoning-2.
\newblock Hugging Face dataset,
  \url{https://huggingface.co/datasets/nvidia/OpenScienceReasoning-2}.

\bibitem[{{OpenAI}(2025)}]{o4mini2025}
{OpenAI}. 2025.
\newblock \href {https://openai.com/index/o3-o4-mini-system-card/} {{OpenAI} o3
  and o4-mini system card}.

\bibitem[{{OpenAI}(2026)}]{gpt55_2026}
{OpenAI}. 2026.
\newblock \href {https://openai.com/index/gpt-5-5-system-card/} {{GPT-5.5}
  system card}.

\bibitem[{Oren et~al.(2024)Oren, Meister, Chatterji, Ladhak, and
  Hashimoto}]{oren2024contamination}
Yonatan Oren, Nicole Meister, Niladri Chatterji, Faisal Ladhak, and
  Tatsunori~B. Hashimoto. 2024.
\newblock Proving test set contamination in black box language models.
\newblock In \emph{International Conference on Learning Representations
  (ICLR)}.
\newblock ArXiv:2310.17623.

\bibitem[{Palen-Michel and Lignos(2023)}]{lrsum2023}
Chester Palen-Michel and Constantine Lignos. 2023.
\newblock Lr-sum: Summarization for less-resourced languages.
\newblock In \emph{Findings of the Association for Computational Linguistics:
  ACL 2023}.
\newblock ArXiv:2212.09674.

\bibitem[{Parmar et~al.(2024)Parmar, Satheesh, Patwary, Shoeybi, and
  Catanzaro}]{parmar2024reuse}
Jupinder Parmar, Sanjev Satheesh, Mostofa Patwary, Mohammad Shoeybi, and Bryan
  Catanzaro. 2024.
\newblock Reuse, don't retrain: A recipe for continued pretraining of language
  models.
\newblock \emph{arXiv preprint arXiv:2407.07263}.

\bibitem[{Penedo et~al.(2024)Penedo, Kydl{\'i}{\v c}ek, Lozhkov, Mitchell,
  Raffel, von Werra, Wolf et~al.}]{fineweb2024}
Guilherme Penedo, Hynek Kydl{\'i}{\v c}ek, Anton Lozhkov, Margaret Mitchell,
  Colin Raffel, Leandro von Werra, Thomas Wolf, et~al. 2024.
\newblock The fineweb datasets: Decanting the web for the finest text data at
  scale.
\newblock In \emph{Advances in Neural Information Processing Systems 37
  (Datasets and Benchmarks Track)}.
\newblock ArXiv:2406.17557.

\bibitem[{Penedo et~al.(2025)Penedo, Kydl{\'i}{\v c}ek, Sabol{\v c}ec, Messmer,
  Foroutan, Kargaran, Raffel, Jaggi, von Werra, and Wolf}]{penedo2025fineweb2}
Guilherme Penedo, Hynek Kydl{\'i}{\v c}ek, Vinko Sabol{\v c}ec, Bettina
  Messmer, Negar Foroutan, Amir~Hossein Kargaran, Colin Raffel, Martin Jaggi,
  Leandro von Werra, and Thomas Wolf. 2025.
\newblock Fineweb2: One pipeline to scale them all -- adapting pre-training
  data processing to every language.
\newblock \emph{arXiv preprint arXiv:2506.20920}.

\bibitem[{Penedo et~al.(2023)Penedo, Malartic, Hesslow, Cojocaru, Cappelli,
  Alobeidli, Pannier, Almazrouei, and Launay}]{refinedweb2023}
Guilherme Penedo, Quentin Malartic, Daniel Hesslow, Ruxandra Cojocaru,
  Alessandro Cappelli, Hamza Alobeidli, Baptiste Pannier, Ebtesam Almazrouei,
  and Julien Launay. 2023.
\newblock The refinedweb dataset for falcon llm: Outperforming curated corpora
  with web data, and web data only.
\newblock In \emph{Advances in Neural Information Processing Systems 36
  (Datasets and Benchmarks Track)}.
\newblock ArXiv:2306.01116.

\bibitem[{Post and Vilar(2018)}]{postvilar2018}
Matt Post and David Vilar. 2018.
\newblock Fast lexically constrained decoding with dynamic beam allocation for
  neural machine translation.
\newblock In \emph{Proceedings of the 2018 Conference of the North American
  Chapter of the Association for Computational Linguistics (NAACL)}.
\newblock ArXiv:1804.06609.

\bibitem[{Rae et~al.(2021)Rae, Borgeaud, Cai, Millican, Hoffmann, Song,
  Aslanides, Henderson, Ring, Young et~al.}]{rae2021gopher}
Jack~W. Rae, Sebastian Borgeaud, Trevor Cai, Katie Millican, Jordan Hoffmann,
  Francis Song, John Aslanides, Sarah Henderson, Roman Ring, Susannah Young,
  et~al. 2021.
\newblock Scaling language models: Methods, analysis \& insights from training
  gopher.
\newblock \emph{arXiv preprint arXiv:2112.11446}.

\bibitem[{Raffel et~al.(2020)Raffel, Shazeer, Roberts, Lee, Narang, Matena,
  Zhou, Li, and Liu}]{raffel2020c4}
Colin Raffel, Noam Shazeer, Adam Roberts, Katherine Lee, Sharan Narang, Michael
  Matena, Yanqi Zhou, Wei Li, and Peter~J. Liu. 2020.
\newblock Exploring the limits of transfer learning with a unified text-to-text
  transformer.
\newblock \emph{Journal of Machine Learning Research}, 21(140):1--67.
\newblock ArXiv:1910.10683.

\bibitem[{Remy et~al.(2024)Remy, Delobelle, Avetisyan, Khabibullina,
  de~Lhoneux, and Demeester}]{remy2024transtok}
Fran{\c{c}}ois Remy, Pieter Delobelle, Hayastan Avetisyan, Alfiya Khabibullina,
  Miryam de~Lhoneux, and Thomas Demeester. 2024.
\newblock Trans-tokenization and cross-lingual vocabulary transfers: Language
  adaptation of llms for low-resource nlp.
\newblock In \emph{Conference on Language Modeling (COLM)}.
\newblock ArXiv:2408.04303.

\bibitem[{Rolnick et~al.(2019)Rolnick, Ahuja, Schwarz, Lillicrap, and
  Wayne}]{rolnick2019replay}
David Rolnick, Arun Ahuja, Jonathan Schwarz, Timothy~P. Lillicrap, and Greg
  Wayne. 2019.
\newblock Experience replay for continual learning.
\newblock In \emph{Advances in Neural Information Processing Systems 32
  (NeurIPS 2019)}.
\newblock ArXiv:1811.11682.

\bibitem[{Romanou et~al.(2025)Romanou, Foroutan, Sotnikova, Chen, Nelaturu,
  Singh, Maheshwary, Altomare, Haggag, Amayuelas et~al.}]{include2025}
Angelika Romanou, Negar Foroutan, Anna Sotnikova, Zeming Chen, Sree~Harsha
  Nelaturu, Shivalika Singh, Rishabh Maheshwary, Micol Altomare, Mohamed~A.
  Haggag, Alfonso Amayuelas, et~al. 2025.
\newblock \href {https://openreview.net/forum?id=k3gCieTXeY} {{INCLUDE}:
  Evaluating multilingual language understanding with regional knowledge}.
\newblock In \emph{The Thirteenth International Conference on Learning
  Representations (ICLR)}.
\newblock ArXiv:2411.19799.

\bibitem[{Rust et~al.(2021)Rust, Pfeiffer, Vuli{\'c}, Ruder, and
  Gurevych}]{rust2021fertility}
Phillip Rust, Jonas Pfeiffer, Ivan Vuli{\'c}, Sebastian Ruder, and Iryna
  Gurevych. 2021.
\newblock How good is your tokenizer? on the monolingual performance of
  multilingual language models.
\newblock In \emph{Proceedings of the 59th Annual Meeting of the Association
  for Computational Linguistics and the 11th International Joint Conference on
  Natural Language Processing (Volume 1: Long Papers)}.
\newblock ArXiv:2012.15613.

\bibitem[{Sainz et~al.(2023)Sainz, Campos, Garc{\'i}a-Ferrero, Etxaniz,
  Lopez~de Lacalle, and Agirre}]{sainz2023contamination}
Oscar Sainz, Jon Campos, Iker Garc{\'i}a-Ferrero, Julen Etxaniz, Oier Lopez~de
  Lacalle, and Eneko Agirre. 2023.
\newblock {NLP} evaluation in trouble: On the need to measure {LLM} data
  contamination for each benchmark.
\newblock In \emph{Findings of the Association for Computational Linguistics:
  EMNLP 2023}.
\newblock ArXiv:2310.18018; anthology 2023.findings-emnlp.722.

\bibitem[{Shi et~al.(2023)Shi, Suzgun, Freitag, Wang, Srivats, Vosoughi, Chung,
  Tay, Ruder, Zhou, Das, and Wei}]{mgsm2023}
Freda Shi, Mirac Suzgun, Markus Freitag, Xuezhi Wang, Suraj Srivats, Soroush
  Vosoughi, Hyung~Won Chung, Yi~Tay, Sebastian Ruder, Denny Zhou, Dipanjan Das,
  and Jason Wei. 2023.
\newblock \href {https://openreview.net/forum?id=fR3wGCk-IXp} {Language models
  are multilingual chain-of-thought reasoners}.
\newblock In \emph{The Eleventh International Conference on Learning
  Representations (ICLR)}.
\newblock ArXiv:2210.03057.

\bibitem[{Silcock et~al.(2023)Silcock, D'Amico-Wong, Yang, and
  Dell}]{silcock2023newsdedup}
Emily Silcock, Luca D'Amico-Wong, Jinglin Yang, and Melissa Dell. 2023.
\newblock Noise-robust de-duplication at scale.
\newblock In \emph{International Conference on Learning Representations
  (ICLR)}.
\newblock ArXiv:2210.04261.

\bibitem[{Singh et~al.(2024{\natexlab{a}})Singh, Romanou, Fourrier, Adelani,
  Ngui, Vila-Suero et~al.}]{globalmmlu2024}
Shivalika Singh, Angelika Romanou, Cl{\'e}mentine Fourrier, David~I. Adelani,
  Jian~Gang Ngui, Daniel Vila-Suero, et~al. 2024{\natexlab{a}}.
\newblock Global mmlu: Understanding and addressing cultural and linguistic
  biases in multilingual evaluation.
\newblock \emph{arXiv preprint arXiv:2412.03304}.

\bibitem[{Singh et~al.(2024{\natexlab{b}})Singh, Vargus, Dsouza, Karlsson,
  Mahendiran, Ko, Shandilya, Patel et~al.}]{ayadataset2024}
Shivalika Singh, Freddie Vargus, Daniel Dsouza, B{\"o}rje~F. Karlsson, Abinaya
  Mahendiran, Wei-Yin Ko, Herumb Shandilya, Jay Patel, et~al.
  2024{\natexlab{b}}.
\newblock \href {https://aclanthology.org/2024.acl-long.620/} {Aya dataset: An
  open-access collection for multilingual instruction tuning}.
\newblock In \emph{Proceedings of the 62nd Annual Meeting of the Association
  for Computational Linguistics (Volume 1: Long Papers)}.
\newblock ArXiv:2402.06619.

\bibitem[{Soboleva et~al.(2023)Soboleva, Al-Khateeb, Myers, Steeves, Hestness,
  and Dey}]{slimpajama2023}
Daria Soboleva, Faisal Al-Khateeb, Robert Myers, Jacob~R. Steeves, Joel
  Hestness, and Nolan Dey. 2023.
\newblock {SlimPajama}: A 627b token cleaned and deduplicated version of
  {RedPajama}.
\newblock Cerebras blog, \url{https://cerebras.ai/blog/slimpajama}.
\newblock Dataset: \texttt{cerebras/SlimPajama-627B} on Hugging Face.

\bibitem[{Soldaini et~al.(2024)Soldaini, Kinney, Bhagia, Schwenk, Atkinson,
  Authur, Bogin, Chandu, Dumas, Elazar et~al.}]{dolma2024}
Luca Soldaini, Rodney Kinney, Akshita Bhagia, Dustin Schwenk, David Atkinson,
  Russell Authur, Ben Bogin, Khyathi Chandu, Jennifer Dumas, Yanai Elazar,
  et~al. 2024.
\newblock Dolma: an open corpus of three trillion tokens for language model
  pretraining research.
\newblock In \emph{Proceedings of the 62nd Annual Meeting of the Association
  for Computational Linguistics (ACL 2024)}.
\newblock ArXiv:2402.00159.

\bibitem[{Tejaswi et~al.(2024)Tejaswi, Gupta, and Choi}]{tejaswi2024}
Atula Tejaswi, Nilesh Gupta, and Eunsol Choi. 2024.
\newblock Exploring design choices for building language-specific llms.
\newblock In \emph{Findings of the Association for Computational Linguistics:
  EMNLP 2024}.
\newblock ArXiv:2405.14670.

\bibitem[{Toshniwal et~al.(2024)Toshniwal, Du, Moshkov, Kisacanin, Ayrapetyan,
  and Gitman}]{openmathinstruct2_2024}
Shubham Toshniwal, Wei Du, Ivan Moshkov, Branislav Kisacanin, Alexan
  Ayrapetyan, and Igor Gitman. 2024.
\newblock Openmathinstruct-2: Accelerating ai for math with massive open-source
  instruction data.
\newblock \emph{arXiv preprint arXiv:2410.01560}.

\bibitem[{{\"U}st{\"u}n et~al.(2024){\"U}st{\"u}n, Aryabumi, Yong, Ko, D'souza
  et~al.}]{aya2024}
Ahmet {\"U}st{\"u}n, Viraat Aryabumi, Zheng-Xin Yong, Wei-Yin Ko, Daniel
  D'souza, et~al. 2024.
\newblock Aya model: An instruction finetuned open-access multilingual language
  model.
\newblock In \emph{Proceedings of the 62nd Annual Meeting of the Association
  for Computational Linguistics}.
\newblock ArXiv:2402.07827.

\bibitem[{Wang et~al.(2024{\natexlab{a}})Wang, Lu, Weber, Ryabinin, Chen, Tang,
  and Stenetorp}]{transwebedu2024}
Jiayi Wang, Yao Lu, Maurice Weber, Max Ryabinin, Yihong Chen, Raphael Tang, and
  Pontus Stenetorp. 2024{\natexlab{a}}.
\newblock Multilingual pretraining using a large corpus machine-translated from
  a single source language.
\newblock \emph{arXiv preprint arXiv:2410.23956}.

\bibitem[{Wang et~al.(2024{\natexlab{b}})Wang, Ma, Zhang, Ni, Chandra, Guo,
  Ren, Arulraj, He, Jiang et~al.}]{mmlupro2024}
Yubo Wang, Xueguang Ma, Ge~Zhang, Yuansheng Ni, Abhranil Chandra, Shiguang Guo,
  Weiming Ren, Aaran Arulraj, Xuan He, Ziyan Jiang, et~al. 2024{\natexlab{b}}.
\newblock {MMLU}-pro: A more robust and challenging multi-task language
  understanding benchmark.
\newblock In \emph{Advances in Neural Information Processing Systems 37
  (NeurIPS 2024), Datasets and Benchmarks Track}.
\newblock ArXiv:2406.01574.

\bibitem[{Xue et~al.(2021)Xue, Constant, Roberts, Kale, Al-Rfou, Siddhant,
  Barua, and Raffel}]{mt5_2021}
Linting Xue, Noah Constant, Adam Roberts, Mihir Kale, Rami Al-Rfou, Aditya
  Siddhant, Aditya Barua, and Colin Raffel. 2021.
\newblock mt5: A massively multilingual pre-trained text-to-text transformer.
\newblock In \emph{Proceedings of the 2021 Conference of the North American
  Chapter of the Association for Computational Linguistics (NAACL)}.
\newblock ArXiv:2010.11934.

\bibitem[{Yamaguchi et~al.(2024)Yamaguchi, Villavicencio, and
  Aletras}]{yamaguchi2024vocab}
Atsuki Yamaguchi, Aline Villavicencio, and Nikolaos Aletras. 2024.
\newblock An empirical study on cross-lingual vocabulary adaptation for
  efficient language model inference.
\newblock In \emph{Findings of the Association for Computational Linguistics:
  EMNLP 2024}.
\newblock ArXiv:2402.10712.

\bibitem[{Zellers et~al.(2019)Zellers, Holtzman, Bisk, Farhadi, and
  Choi}]{hellaswag2019}
Rowan Zellers, Ari Holtzman, Yonatan Bisk, Ali Farhadi, and Yejin Choi. 2019.
\newblock \href {https://doi.org/10.18653/v1/P19-1472} {{H}ella{S}wag: Can a
  machine really finish your sentence?}
\newblock In \emph{Proceedings of the 57th Annual Meeting of the Association
  for Computational Linguistics}, pages 4791--4800.

\bibitem[{Zhao et~al.(2024)Zhao, Zhang, Gao, Zhang, Gui, and
  Huang}]{zhao2024transfer}
Jun Zhao, Zhihao Zhang, Luhui Gao, Qi~Zhang, Tao Gui, and Xuanjing Huang. 2024.
\newblock Llama beyond english: An empirical study on language capability
  transfer.
\newblock \emph{arXiv preprint arXiv:2401.01055}.

\end{thebibliography}

\appendix
\section{Negative results}
\label{app:negatives}
POS tagging regresses under CPT, from
0.18 to 0.01. Exam mathematics is flat at 1.75 points for the released
model, though \fullrun{} reaches 2.75, indicating that data
diversity rather than difficulty may be the binding factor. Instruction-dependent ArmBench
tasks such as judged generation and BLEU-scored QA remain low for all
base-style models including ours. These measure formatting, and we defer
them to an instruction-tuned variant.

\section{Full ArmBench results}
\label{app:armbench}
Table~\ref{tab:armbench-full} reports every ArmBench-LLM task score
produced by the released harness fork for the base model, \newsrun{} at
$3{\times}10^{-5}$, \ourmodel, and \fullrun, including
instruction-dependent tasks (judged generation, BLEU-scored QA) on which
all base-style models score low by construction.

\begin{table*}[t]\centering\footnotesize
\begin{tabular}{lrrrr}
\toprule
ArmBench task (metric) & base E4B & \newsrun{} $3{\times}10^{-5}$ & \textbf{\ourmodel} & \fullrun\\
\midrule
arak (BLEU) & 1.088 & 0.534 & \textbf{1.351} & 1.205\\
belebele (EM) & 0.660 & 0.760 & \textbf{0.900} & 0.820\\
dream (EM) & 0.480 & 0.700 & \textbf{0.840} & 0.800\\
exam\_history (exam pts) & 1.000 & 1.000 & \textbf{2.500} & 0.750\\
exam\_literature (exam pts) & 3.000 & 2.000 & \textbf{3.250} & \textbf{3.250}\\
exam\_math (exam pts) & 1.750 & 0.250 & 1.750 & \textbf{2.750}\\
finer (NER acc) & 0.000 & 0.001 & 0.007 & \textbf{0.036}\\
hartak (EM) & 0.022 & 0.044 & 0.822 & \textbf{0.867}\\
include (EM) & 0.100 & 0.080 & \textbf{0.500} & 0.380\\
mmlu\_pro (MMLU-Pro) & 0.154 & 0.087 & \textbf{0.251} & 0.151\\
ms\_marco (BLEU) & 2.489 & 0.110 & \textbf{3.993} & 3.484\\
paraphrase (BLEU) & 0.575 & 0.322 & \textbf{1.566} & --\\
paraphrase (bert\_score) & 0.748 & 0.663 & \textbf{0.765} & --\\
pioner (NER acc) & 0.010 & 0.007 & \textbf{0.114} & 0.042\\
pos (POS acc) & \textbf{0.180} & 0.000 & 0.010 & 0.000\\
punctuation (acc) & 0.105 & 0.168 & \textbf{0.514} & 0.320\\
scientific (EM) & 0.860 & 0.460 & \textbf{1.000} & \textbf{1.000}\\
sentiment (EM) & 0.150 & 0.150 & 0.470 & \textbf{0.500}\\
short\_sentences\_translation (bert\_score) & 0.491 & 0.621 & \textbf{0.678} & 0.637\\
short\_sentences\_translation (BLEU) & 0.001 & 0.140 & 0.168 & \textbf{0.347}\\
space\_fix (acc) & 0.636 & 0.346 & \textbf{0.718} & 0.632\\
squad (BLEU) & 0.278 & 0.025 & \textbf{0.481} & 0.422\\
syndarin (EM) & 0.040 & 0.360 & \textbf{0.920} & 0.900\\
topic-14class (EM) & 0.004 & 0.043 & \textbf{0.482} & 0.461\\
\bottomrule
\end{tabular}
\caption{Every ArmBench task the fork scores, for the base model,
\newsrun, \ourmodel, and \fullrun{} (\S\ref{sec:results}). Best per row
in bold, ties both. Exam rows are points with negative marking; BLEU rows
are corpus-level; ``--'' marks the one pair (\fullrun{} paraphrase) where
the fork's BERTScore scorer fails reproducibly.}
\label{tab:armbench-full}
\end{table*}

\section{Scale confirmation at 1.3B}
\label{app:ablations}
Table~\ref{tab:13b} confirms the 410M grid of Table~\ref{tab:grid} at 1.3B
parameters. Three recipes were retrained at 27.6B SP tokens each, the 410M ranking
transfers exactly, the union's lead over CulturaX widens from 0.013 to
0.035 mean bpb, and at 1.3B the union overtakes pure \armweb{} even on
the held-out news domain, as larger models exploit data diversity. The table
reports bits-per-byte, lower is better.

\begin{table*}[t]\centering\small
\begin{tabular}{lccccccc}
\toprule
Variant & our-iid & our-tail & FW2-test & hyWiki & LR-Sum & FLORES & Mean\\
\midrule
Union (\armweb+CulturaX) & 0.351 & 0.380 & 0.479 & 0.508 & 0.418 & 0.682 & 0.470\\
CulturaX-hy & 0.404 & 0.448 & 0.500 & 0.537 & 0.437 & 0.705 & 0.505\\
\armweb & 0.367 & 0.405 & 0.575 & 0.649 & 0.460 & 0.815 & 0.545\\
\midrule
\emph{Union @ 410M (ref.)} & 0.426 & 0.433 & 0.545 & 0.603 & 0.468 & 0.717 & 0.532\\
\bottomrule
\end{tabular}
\caption{1.3B confirmation runs (27.6B SP tokens each; bpb, lower is
better), with the 410M union row repeated for reference.}
\label{tab:13b}
\end{table*}

\section{Contamination audit of the perfect MCQA score}
\label{app:audit}
Beyond the pipeline's 13-gram decontamination gates, we audited the perfect
ArmBench scientific-MCQA score (50 of 50) directly. Every benchmark item
was compared against the full \armstem{} training corpus, covering the
Armenian side of both math and science shards with 21.8M distinct 8-grams,
at n-gram lengths 5 and 8
over the normalized signature view. \textbf{At $n{=}8$, zero of the 50 items
share a single 8-gram with the training data.} At $n{=}5$, 15 items share at
least one 5-gram, with a maximum overlap of 5.9\% of an item's 5-grams and a
median of zero, consistent with generic scientific phrasing rather than
item leakage. The base model already scores 0.86 on this 50-item set, and
the adaptation adds seven items. We conclude the score reflects capability, not
memorization, while noting the set's small size in the main text.

\section{Cross-implementation deduplication check}
\label{app:dedupcheck}
We verified the engine twice. First, the reference datasketch library,
run with the identical recipe (word 5-grams, 112 permutations, 14 bands of
8, threshold 0.72) on a 1\% sample of 57{,}801 documents, finds 449
in-sample duplicate documents, of which our full-corpus engine had already
flagged 430 (95.8\%), the remainder being threshold-edge pairs. Second,
we ran NeMo-Curator (v1.1, official container) on the same
post-LID corpus with matched LSH geometry (112 permutations as 14 bands of
8, seed 42) over the same signature view, differing only in shingling
(Curator's 30-character n-grams against our word 5-grams). On the full
5.82M documents Curator flags 23.5\% for removal against our 22.4\%. On a
100K-document sample where both tools ran identically scoped, document-level
decisions agree on 99.2\% of documents and Curator reproduces 71.9\% of
our duplicate clusters exactly. The disagreements are the stochastic LSH
boundary, since clusters both tools find have a median true word-5-gram
Jaccard of 0.79, while clusters only one tool finds have a median of 0.62,
below the 0.72 threshold, where any LSH implementation detects pairs only
probabilistically. We conclude the two implementations are equivalent away
from the threshold boundary and the released corpus is not an artifact of
our engine. Separately, the identical recipe applied to a Russian sister
collection gathered by the same crawler removes only 0.9\% of documents,
against 22.4\% for the Armenian corpus, confirming that the measured
duplication level is a property of the Armenian news ecosystem rather
than of the pipeline.

\section{\armweb{} datasheet summary}
\label{app:datasheet}
The corpus draws on 19 outlets (Table~\ref{tab:outlets}), with the five
largest (News, 1in, Tert, Blognews, 168) contributing 2.71M of the 4.37M
released documents. The released corpus holds 3.3B Gemma
tokens, 1.15B SP tokens, and 789M whitespace words; measured the
same way, CulturaX-hy holds 4.5B Gemma tokens (1.1B words),
HPLT-v2-hy 5.8B (1.4B words), and FineWeb-2-hy
2.3B (558M words), all as distributed.
Publication dates span 1998--2026 (older archives were captured by the
2011--2026 crawl), with a broad plateau of 180K--370K documents per year
from 2012 onward. Median document length is 944 characters and the 90th
percentile is 2{,}784. Western Armenian (\texttt{hyw}) accounts for 4{,}533 documents, with
the remainder Eastern Armenian (\texttt{hye}). Documents carry outlet,
topic, URL, and publication and crawl dates, while author names are withheld. Per-outlet, per-month, and
length distributions ship with the corpus as CSV reports, alongside the
PII scan of the Ethics Statement. The machine-readable leakage-gate
report also ships with the corpus and records zero exact cross-split
collisions, near-duplicate rates of 0.045\%, 0.010\%, and 0.005\% for
validation and the two test sets, and cross-split paragraph overlaps of
30, 63, and 10 shared paragraphs of at least 13 tokens for validation,
test-iid, and the temporal tail.

\begin{table}[t]\centering\small
\setlength{\tabcolsep}{4pt}
\resizebox{\columnwidth}{!}{%
\begin{tabular}{llrr}
\toprule
Outlet & Domain & Documents & Share \\
\midrule
News & \texttt{news.am} & 743{,}867 & 17.0\% \\
1in & \texttt{1in.am} & 683{,}802 & 15.7\% \\
Tert & \texttt{tert.am} & 500{,}341 & 11.5\% \\
Blognews & \texttt{blognews.am} & 405{,}199 & 9.3\% \\
168 & \texttt{168.am} & 376{,}009 & 8.6\% \\
Aravot & \texttt{aravot.am} & 344{,}456 & 7.9\% \\
Armtimes & \texttt{armtimes.com} & 236{,}368 & 5.4\% \\
Lurer & \texttt{lurer.com} & 223{,}624 & 5.1\% \\
Mamul & \texttt{mamul.am} & 218{,}189 & 5.0\% \\
Panarmenian & \texttt{panarmenian.net} & 183{,}936 & 4.2\% \\
Lragir & \texttt{lragir.am} & 177{,}804 & 4.1\% \\
Asekose & \texttt{asekose.am} & 116{,}617 & 2.7\% \\
Radar & \texttt{radar.am} & 42{,}986 & 1.0\% \\
Azatutyun & \texttt{azatutyun.am} & 38{,}727 & 0.9\% \\
Banks & \texttt{banks.am} & 26{,}709 & 0.6\% \\
Iravaban & \texttt{iravaban.net} & 19{,}785 & 0.5\% \\
Armlur & \texttt{armlur.am} & 16{,}243 & 0.4\% \\
Armenpress & \texttt{armenpress.am} & 13{,}613 & 0.3\% \\
Hraparak & \texttt{hraparak.am} & 108 & 0.0\% \\
\midrule
Total & 19 domains & 4{,}368{,}383 & 100\% \\
\bottomrule
\end{tabular}}
\caption{Source outlets of the released \armweb{} corpus with document
counts across all splits. Every document of an outlet resolves to the
single domain shown.}
\label{tab:outlets}
\end{table}

\begin{table}[t]\centering\small
\setlength{\tabcolsep}{4pt}
\resizebox{\columnwidth}{!}{%
\begin{tabular}{lrr}
\toprule
Tokenizer & Tokens/word & Bytes/token \\
\midrule
\armweb{} SP 32K (ours) & 1.48 & 9.56 \\
\armweb{} SP 8K (ours) & 2.09 & 6.77 \\
Gemma-3 / Gemma-4 & 4.15 & 3.42 \\
Qwen3.5 & 5.31 & 2.67 \\
Qwen3 & 7.66 & 1.85 \\
Llama-3.1 & 12.22 & 1.16 \\
\bottomrule
\end{tabular}}
\caption{Armenian tokenizer fertility on \armweb{} validation text (lower
tokens per word is better). The two SentencePiece rows are trained on
\armweb{} and serve the Megatron ablations; the remaining rows are the
stock tokenizers of candidate base models for continued pretraining.}
\label{tab:fertility}
\end{table}

\section{Reproducibility details}
\label{app:repro}
\textbf{CPT training.} Training uses the plain HuggingFace
\texttt{Trainer} under \texttt{torchrun} on 16 nodes of 8 H100s.
Sequences are packed to length 4096 with per-device batch 2 and gradient
accumulation 2, a global batch of 512 sequences or about 2.1M tokens per
step, for 4{,}770 steps or about 10B tokens. The schedule is cosine with
100 warmup steps, the optimizer is AdamW with $\beta_2{=}0.95$ and weight
decay 0.1 in bf16, and data streams through
\texttt{interleave\_datasets} with the probabilities of \S\ref{sec:model}
and seed 42 in all runs. Each run takes about 10 hours of wall-clock time
on 128 H100s, roughly 1{,}250 H100-hours. The replay stream is a
FineWeb-Edu sample and the code stream is Stack-smol. \textbf{ArmSTEM training subset.} The
CPT runs other than \fullrun{} used a 109{,}885-item subset of \armstem{}
(7{,}404 GSM8K, 48{,}584
AceReason-Math, 45{,}883 OpenScience, 8{,}014 OSR-2), about 59M Armenian
and 24M parallel English Gemma tokens, so the 4\%/2\% mixture shares give
$0.04 \times 10\mathrm{B} / 59\mathrm{M} \approx 6.8$ reads of each
Armenian token and $0.02 \times 10\mathrm{B} / 24\mathrm{M} \approx 8.3$
of each English token, the ``7--9 reads'' of \S\ref{sec:model}. A
manifest of the subset ships with the corpus and lists the 104{,}630
training items that appear verbatim in the released corpus. The remaining
5{,}255 OpenScience items were superseded by revised translations before
release and are not in the released corpus. \textbf{Translation models.}
Translator
Gemini-3.1-flash-lite; escalation GPT-5.5; blind re-solver o4-mini;
freeform judge panel: all three, 2/3 majority (accessed June--August
2026). \textbf{Evaluation.} Likelihood suite: LM Evaluation Harness,
zero-shot, \texttt{acc}; ArmBench: the authors' released lighteval fork,
default task configs. ArmenianGPT-1.0-3B is a Mistral-3-based multimodal
checkpoint, evaluated on the likelihood suite through its multimodal
wrapper with text-only inputs and on ArmBench through its extracted text
backbone, which reproduces the wrapper's likelihood scores exactly. The
410M/1.3B grid, repetition, and ladder models are Megatron-LM trainings.
All pipeline code, per-stage reports, and job scripts with their full
configurations are released with the corpora.

\end{document}